\documentclass[lettersize,journal]{IEEEtran}
\usepackage{amsmath,amsfonts}
\usepackage{algorithmic}
\usepackage{algorithm}
\usepackage{array,multirow}
\usepackage[caption=false,font=footnotesize,labelfont=bf,textfont=rm]{subfig}
\usepackage{textcomp}
\usepackage{stfloats}
\usepackage{url}
\usepackage{verbatim}
\usepackage{graphicx}
\usepackage[noadjust]{cite}
\usepackage{balance}
\usepackage[font=footnotesize, labelfont=bf]{caption} 

\usepackage[nolist]{acronym} 
\usepackage{xcolor}
\usepackage{lipsum}
\usepackage{bm}
\usepackage{amsmath,amssymb}

\usepackage[none]{comment}       

\begin{document}
\begin{acronym}[F-VESPA] 
\acro{F-VESPA}{Foot VErtical \& Sagittal Position Algorithm}
\acro{VST}{Variable Stiffness Treadmill}
\acro{VST 2}{Variable Stiffness Treadmill 2}
\acro{ML}{Mediolateral}
\acro{AP}{Anteroposterior}
\acro{VT}{Vertical}
\acro{CoM}{center of mass}
\acro{CoP}{center of pressure}
\acro{GRF}{Ground Reaction Force}
\acro{XcoM}{extrapolated CoM}
\acro{BoS}{base of support}
\acro{TC}{tibia controller}
\acro{AC}{admittance controller}
\acro{LUT}{lookup table}
\acro{FB}{feedback}
\acro{FF}{feedforward}
\acro{AMI}{Average Mutual Information}
\acro{FNN}{False Nearest Neighbor}
\acro{MOS}{margins of stability}
\acro{ROA}{Ruggedized Odyssey Ankle}
\acro{IMU}{inertial measurement unit}
\acro{GC}{gait cycle}
\acro{SP}{stance phase}
\acro{FSM}{finite state machine}
\acro{SSP}{self-selected pace}
\end{acronym}

\title{Control of Powered Ankle-Foot Prostheses on Compliant Terrain: A Quantitative Approach to Stability Enhancement}

\author{Chrysostomos Karakasis,~\IEEEmembership{Graduate Student Member,~IEEE,} Camryn Scully, Robert Salati, Panagiotis Artemiadis,~\IEEEmembership{Senior Member,~IEEE}
\thanks{*This material is based upon work supported by the National Science Foundation under Grants No. \#2020009, \#2015786, \#2025797, and \#2018905. This scientific paper is partially supported by the Onassis Foundation - Scholarship ID: F ZQ029-1/2020-2021.}
\thanks{Chrysostomos Karakasis, Camryn Scully, and Panagiotis Artemiadis are with the Mechanical Engineering Department, at the University of Delaware, Newark, DE 19716, USA. {\tt\small chryskar@udel.edu, cbscully@udel.edu, partem@udel.edu}}%
\thanks{Robert Salati was with the Mechanical Engineering Department, at the University of Delaware, Newark, DE 19716, USA. He is now with the Mechanical Engineering Department, at the Rice University, Houston, TX 77005, USA. {\tt\small rsalati@rice.edu}}%
\thanks{$^*$Corrresponding author: partem@udel.edu}%
}



\maketitle

\begin{abstract}
Walking on compliant terrain presents a substantial challenge for individuals with lower-limb amputation, further elevating their already high risk of falling. While powered ankle-foot prostheses have demonstrated adaptability across speeds and rigid terrains, control strategies optimized for soft or compliant surfaces remain underexplored. This work experimentally validates an admittance-based control strategy that dynamically adjusts the quasi-stiffness of powered prostheses to enhance gait stability on compliant ground. Human subject experiments were conducted with three healthy individuals walking on two bilaterally compliant surfaces with ground stiffness values of $63$ and $25\;\frac{kN}{m}$, representative of real-world soft environments. Controller performance was quantified using phase portraits and two walking stability metrics, offering a direct assessment of fall risk. Compared to a standard phase-variable controller developed for rigid terrain, the proposed admittance controller consistently improved gait stability across all compliant conditions. These results demonstrate the potential of adaptive, stability-aware prosthesis control to reduce fall risk in real-world environments and advance the robustness of human-prosthesis interaction in rehabilitation robotics.
\end{abstract}



\section{Introduction}

\IEEEPARstart{P}{eople} with a lower limb amputation face an increased risk of falling both during and after their clinical recovery compared to able-bodied individuals~\cite{steinberg2019fall,eveld2022factors}. Compliant terrains are an example of surfaces that are commonly found in everyday environments that have been shown to significantly affect walking stability and the risk of falling~\cite{chang2010measures,maclellan2006adaptations,marigold2005adapting}; others include obstacles, uneven surfaces, reduced visibility, etc. Lower-limb prostheses aim to restore mobility and reduce fall risks, with powered prostheses standing out for their energy contribution and adaptability to various tasks and surfaces~\cite{gehlhar2023review,glanzer2018design}. However, their robustness across the varied and unpredictable compliant terrains of daily life remains largely untested.

\begin{figure}[t!]
    \centering
    \includegraphics[width = \linewidth]{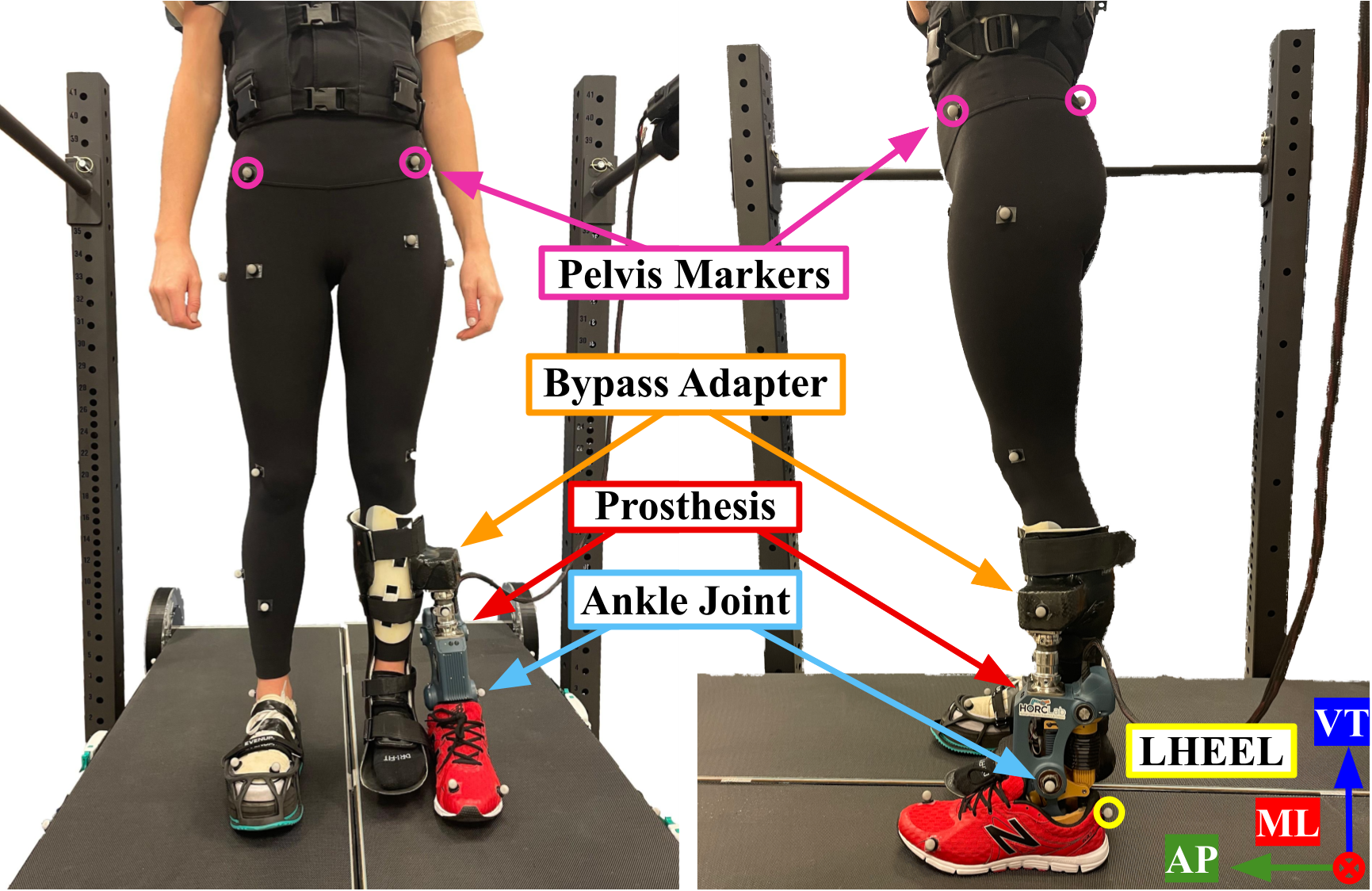}
    \vspace*{-1.5em}
    \caption{Front and side view of a subject standing on the \acf{VST 2}, while wearing the powered ankle-foot prosthesis \acf{ROA} attached to ankle bypass adapter. The magenta circles designate the four markers positioned around the pelvis, specifically, the left and right anterior and posterior superior iliac spine. The yellow circle denotes the marker placed on the left heel (LHEEL) of the prosthetic foot. The red, green, and blue axes correspond to the medial-lateral (ML), anterior-posterior (AP), and vertical (VT) directions, respectively.} 
    \label{fig:subject_prosthesis_vst}
\end{figure}  
Prior research in bipedal locomotion indicates that modifying ankle stiffness may enhance walking stability over compliant surfaces. Initially, humans have been shown to increase leg and ankle stiffness as ground stiffness decreases~\cite{ferris1998running,xie2021compliance}. Mimicking this behavior, a bio-inspired controller improved the robustness of a human-like bipedal model to unilateral ground-stiffness perturbations by increasing leg stiffness with ground compliance~\cite{karakasis2022robust,karakasis2024energy}. In addition to stiffness, the torque-angle relationship, or \textit{quasi-stiffness}~\cite{rouse2012difference}, of the human ankle has been extensively investigated~\cite{lee2016summary,rouse2014estimation}, showing a similar increasing trend with ground compliance~\cite{farley1998mechanism}. Motivated by these findings, we recently showed that increasing the quasi-stiffness of a powered ankle-foot prosthesis improved gait stability during locomotion over a unilaterally compliant terrain~\cite{karakasis2023adjusting}. However, the effect of ankle quasi-stiffness on walking stability has yet to be investigated across different subjects and various bilaterally compliant terrains.

Although quasi-passive ankle-foot prostheses can modulate stiffness with minimal weight and power, they lack the ability to increase push-off work or plantarflexion~\cite{shepherd2017vspa,glanzer2018design,bartlett2021semi}. Powered prostheses address this by generating mechanical power, typically controlled via phase-variable or impedance-based strategies~\cite{gehlhar2023review}. Phase-variable controllers enable continuous adaptation to gait timing~\cite{holgate2009novel,best2021phase,karakasis2023adjusting,cheng2023controlling,best2023data,cheng2024controlling}, while impedance controllers, often combined with finite state machines, are valued for their simplicity and natural interaction~\cite{sup2008design,simon2014configuring,sullivan2023powered,posh2023finite,mohammadi2019variable,kumar2020impedance,lee2024towards,cortino2023data}. However, traditional impedance designs require extensive manual tuning~\cite{reznick2024clinical}, prompting recent work on variable impedance controllers that adjust parameters continuously~\cite{mohammadi2019variable,best2023data,kumar2020impedance,lee2024towards,cortino2023data,reznick2024clinical}. Despite these advances enabling stability over uneven terrains and varying speeds~\cite{best2023data,sullivan2023powered,cheng2024controlling,lee2024towards}, existing controllers remain untested on compliant surfaces.

A key line of research has explored quasi-stiffness modulation in powered prostheses to maintain biomimetic joint behavior~\cite{lenzi2014preliminary,lenzi2014speed,hood2022powered,karakasis2023adjusting,best2023data,gabert2020compact,mendez2020powered,elery2020effects}. Early controllers imposed stiffness profiles from able-bodied data and used \ac{FSM}s to switch between phases, but were limited to level-ground walking~\cite{lenzi2014preliminary,lenzi2014speed,elery2020effects}. More recent efforts introduced variable impedance controllers that continuously adjusted parameters based on gait phase and terrain incline~\cite{best2023data}, yet did not assess stability or performance on compliant surfaces. A phase-based admittance controller was recently proposed to improve stability on compliant terrain~\cite{karakasis2023adjusting}, showing promising results in one subject over unilaterally compliant ground. However, its generalizability across subjects and bilaterally compliant environments remains untested.


While prior work has advanced prosthesis control through quasi-stiffness modulation and phase-based strategies, a critical aspect often underemphasized is gait stability, especially in challenging environments. Gait stability is essential in designing and evaluating lower-limb prosthesis controllers. Traditional assessments focus on trajectory tracking and convergence in phase portraits~\cite{quintero2018continuous,gregg2014virtual,gehlhar2023emulating,best2023data}, but these do not fully reflect user stability. To address this, metrics such as maximum Lyapunov exponents and margins of stability have been widely used across populations and conditions, including people with lower-limb amputation, and walking on compliant terrains~\cite{gates2013frontal,rodrigues2019effects,chang2010measures,maclellan2006adaptations,raffalt2019selection,hof2005condition,mcandrew2011dynamic,bruijn2013assessing,young2012dynamic}. Combining these with phase portraits enables a more comprehensive evaluation of walking stability.

Accurately evaluating gait stability over compliant terrain has been limited by the difficulty of replicating such surfaces in controlled settings. Earlier studies used foam or gym mats with known stiffness~\cite{marigold2005adapting,chang2010measures,maclellan2006adaptations,ferris1998running,xie2021compliance}, but these setups lacked fine stiffness control and made large-scale kinematic and kinetic data collection difficult. To overcome this, our lab developed the \acf{VST}, which could adjust vertical ground stiffness during walking~\cite{barkan2014variable,skidmore2014variable}. While the original \ac{VST} enabled key studies~\cite{skidmore2014investigation,skidmore2015unilateral,drolet2020effects,chambers2022repeated,chambers2022using,karakasis2023adjusting}, it was limited to unilaterally compliant conditions. The newly developed \ac{VST 2} addresses this by independently controlling the stiffness of both belts, enabling controlled studies of bilaterally\footnote{Bilaterally compliant terrains describe surfaces where both legs step on compliant ground.} compliant terrains~\cite{chambers2025variable}.


This paper presents the first experimental evaluation of a novel admittance controller for ankle-foot prostheses aimed at enhancing gait stability over bilaterally compliant terrains. Walking experiments were conducted with three healthy subjects on two compliant surfaces, with stiffness values of $63$ and $25\;\frac{kN}{m}$, representing real-world soft environments. The admittance controller was compared to a standard phase-variable controller designed for rigid ground. The proposed controller improved overall local dynamic stability across the vast majority of conditions. To our knowledge, this is the first study to systematically assess prosthesis control strategies over multiple compliant terrains, addressing a critical gap in real-world prosthetic performance. These findings represent a significant step toward reducing fall risk for individuals with lower-limb amputation navigating soft or unstable surfaces in daily life.

\section{Methods}
\label{section_methods}

\subsection{Ruggedized Odyssey Ankle Prosthesis}
\label{odyssey}
For this study, the powered \acf{ROA} prosthesis was used (Fig.~\ref{fig:subject_prosthesis_vst}), developed by SpringActive, Inc. The \ac{ROA} prosthesis is an updated version of the Walk-Run ankle-foot prosthesis, designed for walking and running~\cite{ward2015rugged,grimmer2017feasibility,naseri2020neuromechanical,clark2022learning}. The ankle consists of a $250\;W$ brushless DC motor, connected in series with a $377\;\frac{kN}{m}$ elastic spring, which then leads to a rigid carbon fiber foot. Therefore, elastic energy is stored and released throughout the gait cycle in the spring, while the motor provides any additional required energy. The prosthesis has two encoders connected to the motor and the ankle joint, and an \ac{IMU} fixed to the inside casing above the ankle. The encoders measure the prosthesis ankle angle and motor position, and the \ac{IMU} measures 3D acceleration and 3D angular velocity. The ankle moment applied by the spring about the ankle joint is estimated through a \ac{LUT} designed by SpringActive, Inc., as a function of the motor position and ankle angle.

\subsection{Tibia Controller}
\label{tibia_controller}

\begin{figure}[!t]
    \centering
    \includegraphics[width=\linewidth]{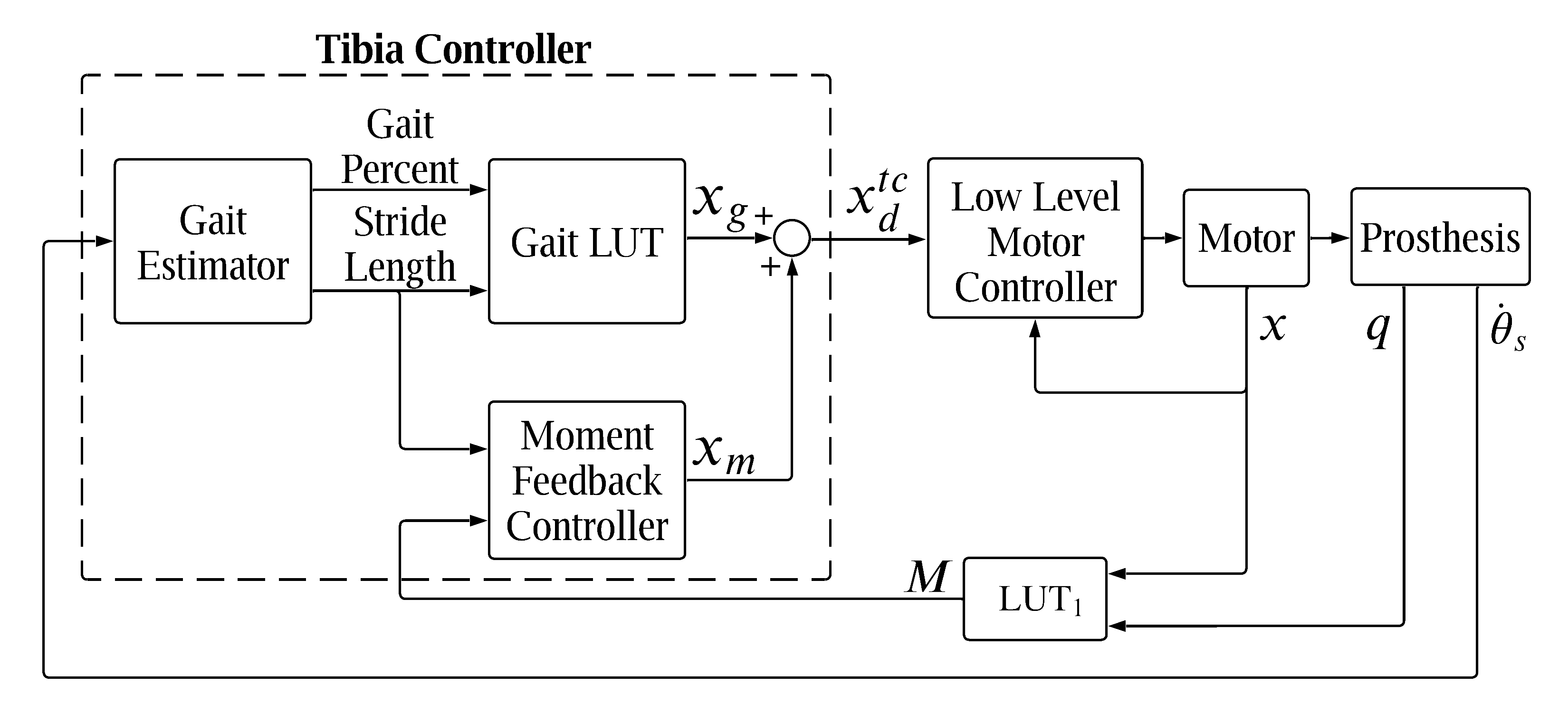}
    \vspace*{-1.5em}
\caption{Block diagram illustrating the implementation of the standard \acf{TC} on the lower-limb prosthesis. The tibia angular velocity $\dot{\theta}_s$, determines the gait percent and stride length, subsequently establishing the reference motor position $x_g$ through a gait \acf{LUT}. The ankle moment $M$ is estimated based on the motor position $x$ and the ankle angle $q$ through a \ac{LUT} ($\text{LUT}_{1}$). A moment feedback controller exports the motor command $x_m$ based on the ankle moment $M$ to enhance the reference motor command at low stride lengths. The two motor commands are combined to yield the final motor position command $x_{d}^{tc}$, sent to the low-level motor controller. The low-level motor controller directs the rotation of the prosthesis ankle joint by issuing precise position commands to the motor.}
    \label{fig:diagram_tibia}
\end{figure}

The \ac{ROA} prosthesis is by default controlled using the \ac{TC}, a continuous phase-variable controller proposed in~\cite{holgate2009novel}. In essence, the \ac{TC} leverages tibia kinematics to generate a reference motor trajectory, mimicking ankle angle trajectories observed in the gait of healthy non-disabled individuals. A block diagram illustrating the implementation of the \ac{TC} to the \ac{ROA} prosthesis is shown in Fig.~\ref{fig:diagram_tibia}. Initially, the tibia angular velocity $\dot{\theta}_{s}$ is calculated through the measurements of a gyroscope included in the \ac{IMU} of the prosthesis. Then, the gait percent and stride length are determined using the phase plane methodology outlined in~\cite{holgate2009novel}. By combining the gait percent and stride length, an appropriate reference ankle angle is derived using a gait \ac{LUT}, which fits a surface function obtained from motion capture data of healthy non-disabled humans. Consequently, the prosthesis kinematics are used to calculate a corresponding motor position $x_{g}$, providing a continuous reference motor trajectory across the entire gait cycle. 

In parallel, a moment feedback controller outputs another motor command $x_m$ based on the ankle moment $M$, which is again computed based on the motor position $x$ and the ankle angle $q$ through a \ac{LUT} ($\text{LUT}_{1}$):
\begin{equation}
    M = LUT_{1}(x,q).
    \label{eq:LUT1}
\end{equation}
The moment feedback controller is an additional component to the original \ac{TC} described in~\cite{holgate2009novel}, which SpringActive added to amplify the motor command of the \ac{TC} at low stride lengths. The final motor output $x_{d}^{tc}$ is a weighted combination of the $x_g$ and $x_m$ commands, which prioritizes the moment feedback controller at low stride lengths ($L_{s}^{n}<1$) by dynamically adjusting each weight:
\begin{equation}
    x_{d}^{tc} = 
        \begin{cases}
            a x_{m} + (1-a)x_{g}, & \text{if } L_{s}^{n}<1,\\
            x_{g}, & \text{else,}
        \end{cases}
    \label{general_admittance}
\end{equation}
\begin{equation}
    \text{where, }a = 0.5\cos{(\pi L_{s}^{n})}+0.5,
    \label{general_admittance}
\end{equation}
\begin{equation}
    L_{s}^{n} = \frac{L_{s}}{SSP},
    \label{general_admittance}
\end{equation}
where $L_{s}$ is the stride length and $L_{s}^{n}$ is the stride length normalized to \ac{SSP}; i.e.\ $L_{s}^{n} = 1$ corresponds to \ac{SSP}.
Ultimately, the low-level motor controller, developed by SpringActive, functions as a finely tuned proportional-derivative feedback controller that guarantees accurate motor position tracking across all anticipated speeds and loads associated with the \ac{ROA} device.


The \ac{TC} has been successfully used by individuals with transtibial amputation for walking and running on rigid terrain~\cite{ward2015rugged,holgate2009novel}, and was recently validated for walking over unilaterally compliant surfaces~\cite{karakasis2023adjusting}. However, a key limitation of the \ac{TC} is its inability to adapt the desired ankle trajectory in response to changes in ground compliance. This lack of adaptability can result in unstable gait patterns and increased risk of falls, particularly during abrupt transitions from rigid to soft surfaces -- conditions commonly encountered in daily life. This limitation highlights the need for a more robust control approach capable of maintaining stability across varying terrain compliance.

\subsection{Admittance Controller}
\label{admittance_controller}

Inspired by the increase of leg and ankle stiffness observed in humans during locomotion over compliant terrains~\cite {farley1998mechanism,ferris1998running,xie2021compliance}, the \ac{AC} was introduced in~\cite{karakasis2023adjusting}. In short, the \ac{AC} allows the adjustment of the ankle quasi-stiffness in powered prosthetic devices, aiming to assist in walking over compliant terrains. A brief overview of the \ac{AC} is presented below.

The \ac{AC} builds upon the \ac{TC} analyzed in Subsection~\ref{tibia_controller}, extended to permit the modification of ankle quasi-stiffness as required. In general, the \ac{AC} imposes the following spring-mass-damper behavior on the ankle joint of one degree of freedom:
\begin{equation}
    M=K_d(q_d-q_e)+B_d(\dot{q}_d-\dot{q}_e)+I_d(\ddot{q}_d-\ddot{q}_e),
    \label{general_admittance}
\end{equation}
where $M$ is the externally applied moment, $K_d, B_d, I_d$ are the stiffness, damping, and inertia of the admittance controller, $q_d, \dot{q}_d, \ddot{q}_d$ are the desired ankle angle and its time derivatives, and $q_e, \dot{q}_e, \ddot{q}_e$ are the equilibrium ankle angle and its time derivatives. As only the control of the ankle quasi-stiffness was of interest in this work, the damping and inertia were set to zero ($B_d=I_d=0$), yielding the following feedback control law:
\begin{equation}
    q_d=q_e+\frac{M}{K_d}.
    \label{eq:final_admittance}
\end{equation}
As a consequence, the adjustment of the ankle quasi-stiffness is achieved by specifying a value for the desired stiffness $K_d$ of the admittance controller in~\eqref{eq:final_admittance}.

\begin{figure}[!t]
    \centering
    \includegraphics[width=\linewidth]{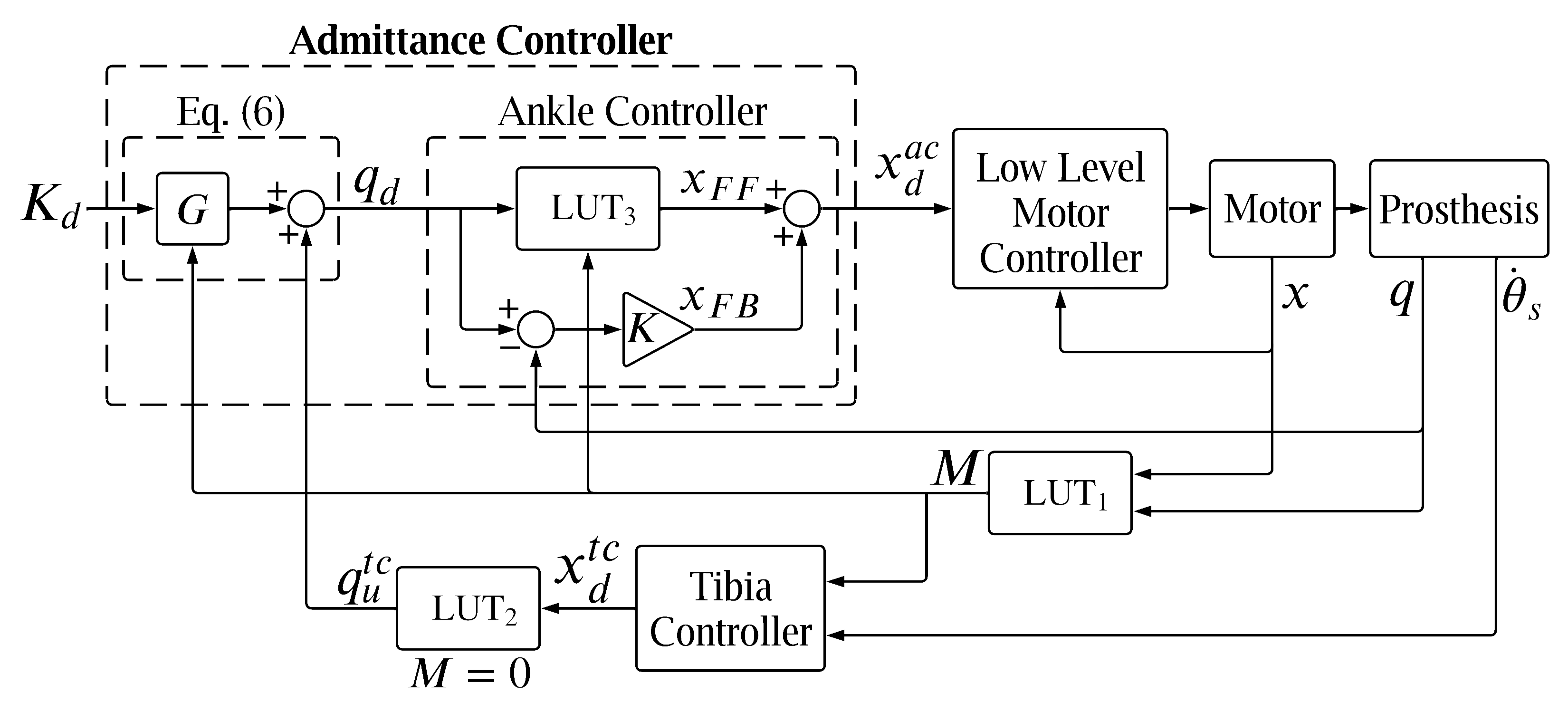}
    \vspace*{-1.5em}
\caption{Block diagram illustrating the implementation of the proposed \acf{AC} on the lower-limb prosthesis. The desired stiffness $K_d$ is set, and the admittance controller calculates an ankle angle offset based on the applied moment $M$. Again, $M$ is derived from a \ac{LUT} based on the motor position $x$ and the ankle angle $q$ ($\text{LUT}_{1}$). Simultaneously, the tibia controller calculates a motor position command $x_{d}^{tc}$ based on the ankle moment and the tibia angular velocity $\dot{\theta}_s$. The virtual unloaded reference ankle angle $q_{u}^{tc}$ is calculated through an inverted version of the ankle moment \ac{LUT} ($\text{LUT}_{2}$), based on $x_{d}^{tc}$ and a virtual zero ankle moment. The ankle angle offset is added to $q_{u}^{tc}$, producing a desired ankle angle $q_d$, which is then fed into the ankle controller. The ankle controller consists of a \acf{FF} and a proportional \acf{FB} system, which combined together yield the final motor position command $x_{d}^{ac}$ sent to the low-level motor controller. The low-level motor controller sends position commands to the motor, causing the rotation of the prosthesis ankle joint.}
    \label{fig:diagram_adm}
\end{figure}

The implementation of the \ac{AC} to the \ac{ROA} prosthesis is illustrated in the block diagram of Fig.~\ref{fig:diagram_adm}. Initially, the user selects a desired stiffness $K_{d}$, and the externally applied moment $M$ is estimated using the ankle moment calculated by the prosthesis ($\text{LUT}_{1}$), as shown in \eqref{eq:LUT1}. Subsequently, the function $G=\frac{M}{K_d}$ outputs an ankle angle offset based on the $K_d$ and $M$. Following~\eqref{eq:final_admittance}, this offset is added to the equilibrium ankle angle $q_e$, resulting in the desired ankle angle $q_d$. As the equilibrium ankle angle, we used the virtual unloaded reference ankle angle $q_{u}^{tc}$, which is calculated through an inverted version of the ankle moment \ac{LUT} ($\text{LUT}_{2}$), based on the reference motor position $x_{d}^{tc}$ of the \ac{TC} and a virtual zero ankle moment:
\begin{equation}
    q_{e} = q_{u}^{tc} = LUT_{2}(x_{d}^{tc},M=0).
\end{equation}
This enables the \ac{AC} to be decoupled from the intrinsic stiffness of the prosthesis, imposed by its dynamics, while inheriting the robust and physiological behavior of the \ac{TC}. Up to this point, the proposed \ac{AC} takes a desired stiffness command $K_{d}$ as input and generates a desired ankle angle $q_{d}$ to emulate the target quasi-stiffness. Therefore, the \ac{AC} imposes a desired constant stiffness $K_{d}$ around an equilibrium angle $q_{e}$, derived from an unloaded version of the \ac{TC}~\eqref{eq:final_admittance}.

Next, the desired ankle angle is transformed into a motor position command using a two-component controller (Ankle Controller). First, the dynamics of the prosthesis are fitted in a \acf{FF} system that uses a \ac{LUT} ($\text{LUT}_{3}$) to compute the desired motor position $x_{FF}$ as a function of the desired ankle angle $q_d$ and the measured ankle moment $M$:
\begin{equation}
    x_{FF} = LUT_{3}(q_{d},M).
\end{equation}
The $\text{LUT}_{3}$ was obtained by inverting the \ac{LUT} created by SpringActive ($\text{LUT}_{1}$) that approximates the ankle moment $M$ as a function of the motor position $x$ and the ankle angle $q$~\eqref{eq:LUT1}. Second, a proportional \acf{FB} controller with a gain $K=0.45$ produces another motor position command $x_{FB}$ to minimize discrepancies between the desired and actual ankle angles:
\begin{equation}
    x_{FB} = K(q_{d}-q).
\end{equation}
Then, the feedback and feedforward motor position commands are combined together to yield the desired motor position $x_{d}^{ac}$:
\begin{equation}
    x_{d}^{ac} = x_{FF}+x_{FB}.
\end{equation}
Finally, the desired motor position is forwarded to the low-level motor controller, which ensures precise motor position tracking across conditions, as described for the \ac{TC} (Subsection~\ref{tibia_controller}).
\subsection{Experimental Protocol}
\label{exp_protocol}
For this study, three healthy non-disabled human subjects (two female and one male, age: $22\pm1.73$, height: $1.66\pm0.03\;m$, weight: $58.98\pm2.27\;kg$) completed treadmill walking trials with the \ac{ROA} prosthesis on a unique instrumented treadmill, the \acf{VST 2}~\cite{chambers2025variable}. Similar to previous studies~\cite{clark2022learning,karakasis2023adjusting,posh2023finite}, the prosthesis was attached to the subjects through a carbon fiber ankle bypass adapter fitted around their left shank, without restraining knee movement (see Fig.~\ref{fig:subject_prosthesis_vst}). To address the height difference introduced by the bypass adapter, an adjustable shoe-lift was worn below the right shoe (Evenup Corp., Buford, GA).

The subjects did not have any prior experience with the prosthesis and only performed two visits, one for training and one for experimentation. On the first visit, the subjects first walked on the \ac{VST 2} at a self-selected speed without the prosthesis in order to get accustomed to the different stiffness levels of the treadmill. Next, the subjects walked with the prosthesis at a rigid stiffness level for various five-minute trials, where the treadmill speed was gradually increased until the subjects had identified their comfortable speed. The prosthesis was controlled using the \ac{TC} for the aforementioned trials. After a comfortable walking speed had been identified, the subjects completed short walking trials at that speed over rigid and bilaterally compliant ground stiffness levels, using both the \ac{TC} and the \ac{AC} at three different desired stiffness levels ($K_d=10$, $15$, and $20$). \P{The \ac{AC} $K_{d}$ values were chosen to align with ankle stiffness values reported in both human and prosthetic studies \cite{rouse2014estimation,farley1998mechanism,xie2021compliance,clites2021understanding}.}

On the second visit, the subjects completed eight walking trials at a self-selected walking speed, which all subjects independently chose to be $0.65\;\frac{m}{s}$. For the first trial, the subjects walked with the standard \ac{TC} over a rigid stiffness level for 300 gait cycles. For the next two trials, the subjects walked with the standard \ac{TC} for 200 gait cycles over two bilaterally compliant terrains of 63 and $25\;\frac{kN}{m}$. For the remaining six trials, the prosthesis was controlled using the \ac{AC} at the following desired stiffness $K_{d}$ values: 10, 15, and $20\;\frac{Nm}{deg}$, over the two bilaterally compliant terrains of 63 and $25\;\frac{kN}{m}$. As a reference, these two levels of ground stiffness correspond to surfaces similar to rubber on foam and a foam pad~\cite{bosworth2016robot}. Subjects were given a five-minute rest between trials to mitigate fatigue. During all trials, a body weight support harness was attached around the torso of the subjects to ensure safety without offloading any weight during normal gait. Additionally, handrails were positioned on the sides of the treadmill for added safety, although they remained unused (Fig.~\ref{fig:subject_prosthesis_vst}). Informed consent was given, and the presented experimental protocol was approved by the University of Delaware Institutional Review Board (IRB ID\#: 1520622-8). A supplemental video of all subjects walking across these conditions is accessible for download.

A motion capture system equipped with eight cameras was used to collect kinematic data at $100\;Hz$ (Vicon Motion Systems Ltd.). Specifically, 22 reflective markers were positioned on the lower body of the subjects and their 3D positions were tracked along the \acf{ML}, \acf{AP}, and \acf{VT} directions (Fig.~\ref{fig:subject_prosthesis_vst}). In this study, focus was given only to the markers placed around the pelvis, as well as the heel (LHEEL) of the left prosthetic foot (see Fig.~\ref{fig:subject_prosthesis_vst}). The vertical \ac{GRF} response and the \ac{CoP} under each leg of the subjects was recorded at $65\;Hz$ using a force sensor mat (\textit{Medical Sensor 3140\textsuperscript{TM}}, Tekscan, Inc., South Boston, MA) placed underneath both belts of the treadmill. The recorded force mat data were first upsampled from $65\;Hz$ to $100\;Hz$ using linear interpolation to match the frequency rate of the motion capture data, and then were filtered using a moving average filter with a window of 10 samples. Data from the prosthesis, such as ankle angle and moment, were recorded at a rate of $100\;Hz$. To mitigate any transient artifacts, the first 25 out of the total 200 recorded strides were excluded in all trials.
\subsection{Stability Measures}
\label{stability_measures}
This section presents the stability metrics used to assess gait and estimate the subjects' risk of falling during the walking trials. Similar to previous works~\cite{young2012dynamic,havens2018analysis}, the position of the \ac{CoM} $\bm{p_{CoM}}(t)$ with respect to time $t$ for each subject was estimated as the mean 3D position of the left and right anterior ($\bm{p_{LASI}}(t)$ and $\bm{p_{RASI}}(t)$) and posterior superior ($\bm{p_{LPSI}}(t)$ and $\bm{p_{RPSI}}(t)$) iliac spine markers highlighted in Fig.~\ref{fig:subject_prosthesis_vst}:

\begin{equation}
\bm{p_{CoM}}(t) = \frac{1}{4} \sum_{m} \bm{p_{m}}(t), 
\end{equation}
where $m \in \{ \text{LASI}, \text{RASI}, \text{LPSI}, \text{RPSI} \}$.
The \ac{CoM} velocity for each subject was computed as the time derivative of the \ac{CoM} position:
\begin{equation}
    \bm{\dot{p}_{CoM}}(t)  = \frac{\Delta\bm{p_{CoM}}(t)}{\Delta t},
\end{equation}
where $\Delta t = 0.01\;s$.
\subsubsection{Maximum Lyapunov Exponents}
\label{max_lyap_exponent}

As outlined in~\cite{rosenstein1993practical}, the maximum Lyapunov or divergence exponents quantify a system's sensitivity to small perturbations, by tracking the exponential rate of divergence between neighboring trajectories in state space. In particular, larger exponents signify heightened sensitivity to local perturbations, indicating greater divergence and reduced local dynamic stability. Following the approach reported in previous studies, we employed the maximum Lyapunov exponents to evaluate the local dynamic stability of human walking~\cite{bruijn2013assessing}.

The maximum Lyapunov exponents were calculated using the algorithm proposed by Rosenstein in~\cite{rosenstein1993practical}, following the approach outlined in~\cite{terrier2013non}. First, the estimated 3D \ac{CoM} position was filtered via a $2^{nd}$ order, low-pass Butterworth filter (cut-off frequency of $10\;Hz$), to remove any artifacts unrelated to musculoskeletal motion~\cite{england2007influence}. Then, the 3D \ac{CoM} velocity $\bm{\dot{p}_{CoM}}(t)$ was derived, consisting of a \ac{ML}, \ac{AP}, and a \ac{VT} component (Fig.~\ref{fig:subject_prosthesis_vst}). Following the guidelines reported in~\cite{bruijn2009statistical,raffalt2019selection}, the velocity time series were time-normalized to a fixed number of 150 strides and a fixed number of 15000 data points per time series, to obtain accurate estimates of the divergence exponents. From each time-normalized velocity $\bm{\dot{p}_{\alpha}}(t_{n}), \alpha \in \{\text{ML, AP, VT}\}$, a high-dimensional attractor $\bm{s_{\alpha}}(t_{n})$ was recreated through the method of delay embedding~\cite{takens2006detecting}:
\begin{equation}
    \bm{s_{\alpha}}(t_{n}) = 
    \left[\bm{\dot{p}_{\alpha}}(t_{n}) \; \bm{\dot{p}_{\alpha}}(t_{n}+\tau_{\alpha}) \; \dots \; \bm{\dot{p}_{\alpha}}(t_{n}+(d_{\alpha}-1)\tau_{\alpha})\right],
\end{equation}
with $\tau_{\alpha}$ and $d_{\alpha}$ denoting the time delay and embedding dimensions, respectively. Time delays and embedding dimensions were computed separately for each one of the three velocity types (\ac{ML}, \ac{AP}, and \ac{VT}) using the \ac{AMI} algorithm~\cite{kantz2004nonlinear}, and the \ac{FNN} algorithm~\cite{rhodes1997false}, respectively. The average time delays and embedding dimensions across all trials are reported in Table~\ref{table:time_delays_emb_dimensions}.
\begin{table}[t!]
    \centering
    \caption{Average Time Delays and Embedding Dimensions}
    \begin{tabular}{|c|c|c|}
        \hline
        \textbf{Velocity Type} & \textbf{Time Delay $\tau$ (samples)} & \textbf{Embedding Dimension} $d$ \\
        \hline
        \ac{ML} & $10 \pm 0.04$ & $3.27 \pm 0.45$ \\
        \hline
        \ac{AP} & $6.95 \pm 0.23$ & $4 \pm 0.04$ \\
        \hline
        \ac{VT} & $7.39 \pm 1.46$ & $4$ \\
        \hline
    \end{tabular}
    \label{table:time_delays_emb_dimensions}
\end{table}

\begin{figure}[!t]
    \centering
    \includegraphics[width=\linewidth]{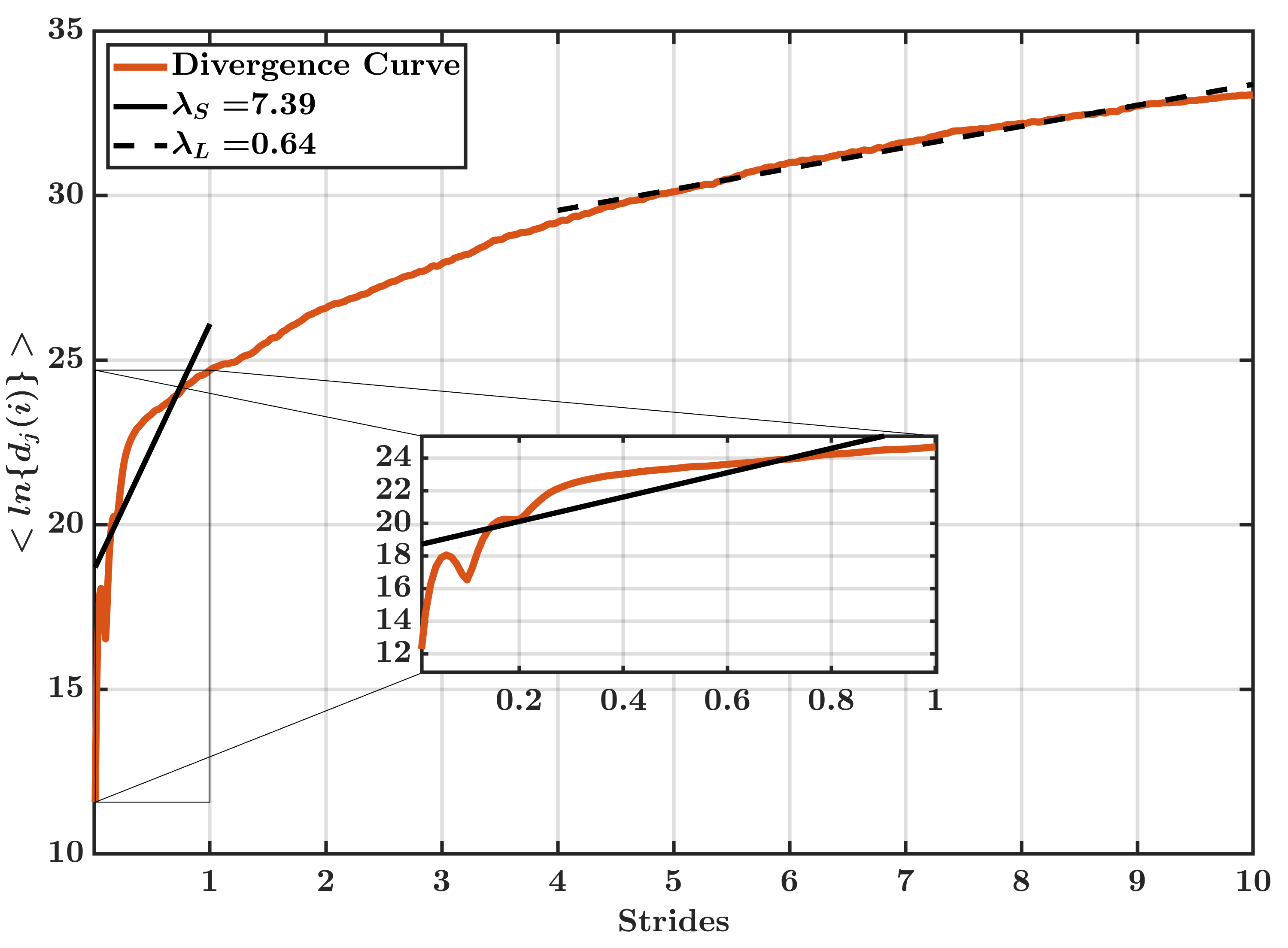}
    \vspace*{-1.5em}
    \caption{Representative divergence curve from the \ac{ML} velocity signal for the trial of the first subject with the \acf{TC} over rigid terrain. The orange line represents the divergence curve, while the black solid and dashed lines indicate the least-square fits over the intervals of 0-1 and 4-10 strides, respectively. Divergence exponents $\lambda_{S}$ and $\lambda_{L}$ represent the slopes of the least-square fits, expressed as $\left<\ln\{d_{j}(i)\}\right>/\text{stride}$. Inset: Divergence curve over the interval of 0-1 strides.} 
    \label{fig:divergence_curves}
\end{figure}

For every point within each reconstructed state space $\bm{s_{\alpha}}(t_{n})$, we identified the nearest neighbor $j$ and tracked the Euclidean distance between them over a 10-stride interval, yielding a time-distance curve $d_{j}(i)$, with $i$ denoting discrete time instances. The natural logarithmic transformation was applied to all time-distance curves across all neighbor pairs $j$, and subsequently, an average was computed to obtain the divergence curve, denoted as $\left<\ln\{d_{j}(i)\}\right>$. As in previous works, the short-term ($\lambda_{S}$) and long-term ($\lambda_{L}$) Lyapunov exponents were defined as the slopes of linear least-square fits applied to each divergence curve during the intervals of 0-1 strides and 4-10 strides, respectively: 
\begin{equation}
    \lambda_S = \frac{ \left< \ln \left\{ d_{j} \left( i_{s_{1}} \right) \right\} \right> - \left< \ln \left\{ d_{j} \left( i_{s_{0}} \right) \right\} \right> }{ 1 - 0 },
\end{equation}
\begin{equation}
    \lambda_L = \frac{ \left< \ln \left\{ d_{j} \left( i_{s_{10}} \right) \right\} \right> - \left< \ln \left\{ d_{j} \left( i_{s_{4}} \right) \right\} \right> }{ 10 - 4 },
\end{equation}
where $i_{s_{0}},i_{s_{1}},i_{s_{4}},i_{s_{10}}$ denote the time instances corresponding to strides 0, 1, 4, and 10, respectively.
The mean duration of strides across all trials was $1.47 \pm 0.06\;s$. To account for variability within each trial, divergence exponents were computed for 25 overlapping windows of 150 strides (175 strides in total) in each trial, and average values along with standard deviations were determined for each exponent~\cite{bruijn2009statistical}. As a representative example, a divergence curve derived from the \ac{VT} velocity signal for the trial with the tibia controller is illustrated in Fig. \ref{fig:divergence_curves}.
\subsubsection{Margins of Stability}
\label{margins_stability}

The \ac{MOS} have been adopted as a measure of stability, as they depend on the distance between the \ac{XcoM} and the \ac{BoS}~\cite{hof2005condition}. The \ac{XcoM} represents the projected future position of the \ac{CoM} extrapolated from its velocity, while the \ac{BoS} signifies the position of either foot during walking. Intuitively, during unstable walking and ultimately falling, the \ac{CoM} approaches and eventually exceeds the \ac{BoS}, showing that greater positive \ac{MOS} values indicate increased stability. Therefore, the margins of stability, influenced by foot placement, play a crucial role in maintaining stability during walking by ensuring that the feet are positioned laterally and anterior to the \ac{XcoM}, allowing for effective compensation in response to perturbations and contributing to overall stability~\cite{bruijn2013assessing}. As introduced in~\cite{hof2005condition}, the \ac{XcoM} was defined as 
\begin{equation}
    \bm{XcoM}(t) = \bm{p_{CoM}}(t) + \frac{\bm{\dot{p}_{CoM}}(t)}{\omega_{o}},\quad \omega_{o} = \sqrt{\frac{g}{l}}
\end{equation}
where $\bm{p_{CoM}}(t)$ and $\bm{\dot{p}_{CoM}}(t)$ are the 3D \ac{CoM} position and velocity, respectively, and $\omega_{o}$ is the eigenfrequency of a hanging non-inverted pendulum of length $l$, while $g=9.81\;\frac{m}{s^2}$ is the gravity acceleration. Similar to \cite{young2012dynamic}, the average Euclidean distance between the \ac{CoM} and the LHEEL marker was used to calculate the length $l$, evaluated at foot-strike ($l = 0.97\pm 0.04\;m$). The foot-strike events were determined using the real-time kinematic algorithm \acs{F-VESPA}~\cite{karakasis2021f,karakasis2021real}.

\begin{figure}[!t]
    \centering
    \includegraphics[width=\linewidth]{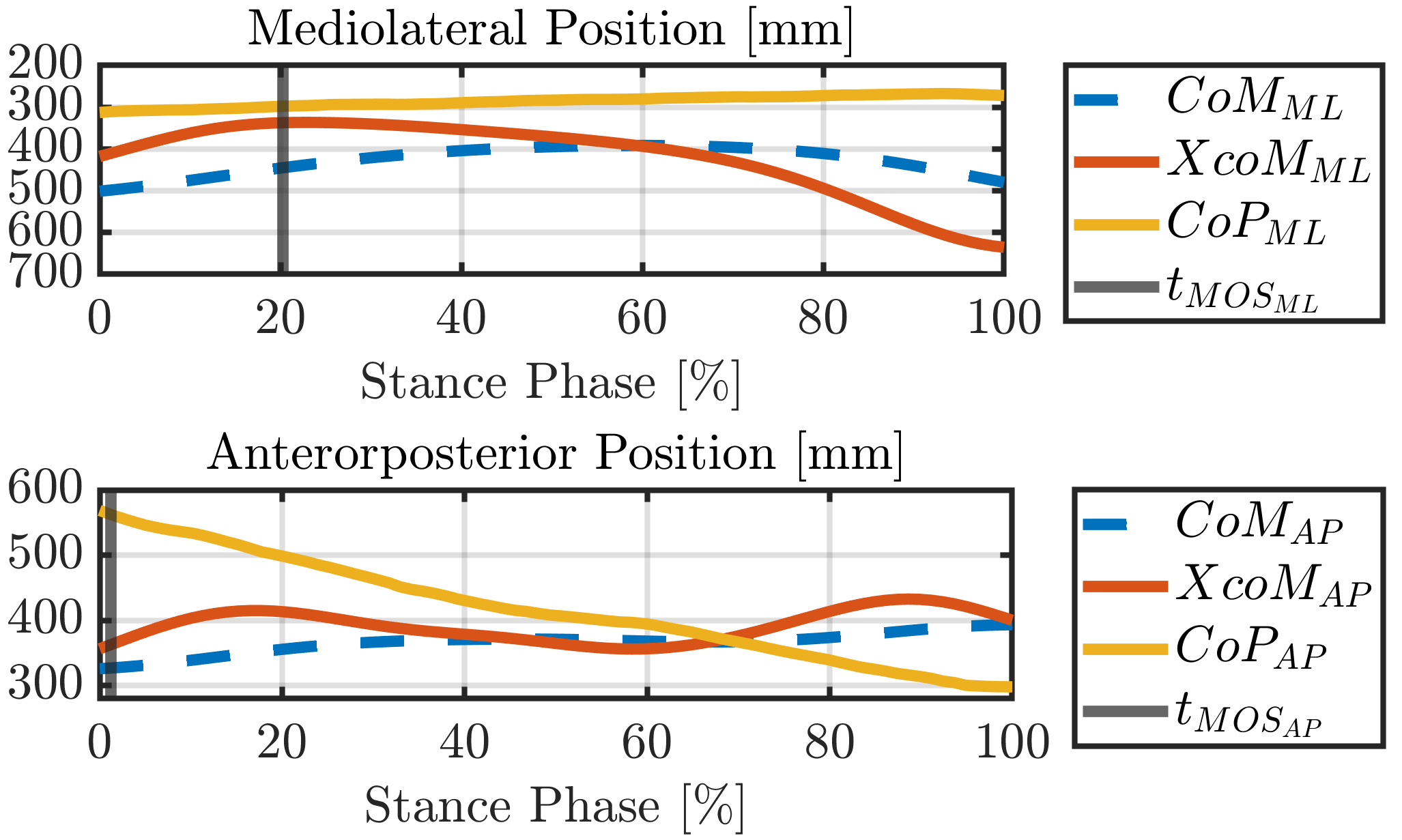}
    \vspace*{-1.5em}
    \caption{Representative example for the calculation of the mediolateral $\text{MOS}_{\text{ML}}$ (top) and anteroposterior $\text{MOS}_{\text{AP}}$ (bottom) margins of stability for the right side (intact limb) during one gait cycle. Blue dashed lines denote the filtered position of the \ac{CoM}, while solid orange and yellow lines depict the filtered positions of the \ac{XcoM} and the \ac{CoP}, respectively. Vertical solid black lines indicate the timings during the stance phase at which the margins of stability were derived; when the distance between the \ac{CoP} and the \ac{XcoM} was minimized in the mediolateral (top) and maximized in the anteroposterior (bottom) direction, respectively. For this example, $\text{MOS}_{\text{ML}} = 39.21\;mm$ and $\text{MOS}_{\text{AP}} = 212.06\;mm$. For brevity, only the derivation for the right side (intact limb) is shown in this figure, with the left side (prosthetic limb) following a similar procedure.} 
    \label{fig:MOS_example}
\end{figure}

The left and right mediolateral margins of stability ($\text{MOS}_{\text{ML}}^{L}$ and $\text{MOS}_{\text{ML}}^{R}$) were determined by calculating the minimum medial-lateral distance, within each gait cycle $k$, between the \ac{XcoM} and the left and right lateral \ac{BoS}, defined by the left and right \ac{CoP} ($\text{CoP}^{\text{L}}_{\text{ML}}$ and $\text{CoP}^{\text{R}}_{\text{ML}}$), respectively~\cite{hof2007control,curtze2024notes}:
\begin{equation}
    \text{MOS}^{\text{L}}_{\text{ML}}(k) = \min_{i}{\left(\left\lVert\text{CoP}^{\text{L}}_{\text{ML}}(i) - \bm{XcoM}_{\text{ML}}(i)\right\rVert\right)},
\end{equation}
\begin{equation}
    \text{MOS}^{\text{R}}_{\text{ML}}(k) = \min_{i}{\left(\left\lVert\text{CoP}^{\text{R}}_{\text{ML}}(i) - \bm{XcoM}_{\text{ML}}(i)\right\rVert\right)},
\end{equation}
with $i$ denoting discrete time instances within each gait cycle.
Similarly, the left and right anterior-posterior margins of stability ($\text{MOS}_{\text{AP}}^{L}$ and $\text{MOS}_{\text{AP}}^{R}$) were defined by calculating the maximum anterior-posterior distance during each gait cycle $k$ between the \ac{XcoM} and the left and right anterior \ac{BoS}, defined by the left and right \ac{CoP} ($\text{CoP}^{\text{L}}_{\text{AP}}$ and $\text{CoP}^{\text{R}}_{\text{AP}}$), respectively~\cite{curtze2024notes}:
\begin{equation}
    \text{MOS}^{\text{L}}_{\text{AP}}(k) = \max_{i}{\left(\left\lVert\text{CoP}^{\text{L}}_{\text{AP}}(i) - \bm{XcoM}_{\text{AP}}(i)\right\rVert\right)},
\end{equation}
\begin{equation}
    \text{MOS}^{\text{R}}_{\text{AP}}(k) = \max_{i}{\left(\left\lVert\text{CoP}^{\text{R}}_{\text{AP}}(i) - \bm{XcoM}_{\text{AP}}(i)\right\rVert\right)},
\end{equation}
with $i$ again indicating discrete time instances within each gait cycle.
Kinematic data were filtered via a $4^{th}$ order, low-pass Butterworth filter (cut-off frequency of $5\;Hz$), while the \ac{CoP} data were filtered using a moving average with a window of 10 samples. A representative example for the calculation of the mediolateral $\text{MOS}_{\text{ML}}$ (top) and anteroposterior $\text{MOS}_{\text{AP}}$ (bottom) margins of stability is shown in Fig.~\ref{fig:MOS_example}. Average values and standard deviations were computed for each left and right \ac{MOS} over a total of 175 steps across all trials.
\section{Results}
\label{results}
This section involves a comprehensive evaluation of both the admittance and tibia controllers over two bilaterally compliant terrains, in a total of eight walking trials for each of the three subjects. In the first two trials, the tibia controller was tested over two compliant terrains of 63 and $25\;\frac{kN}{m}$. For the last six trials, the admittance controller was tested using the following desired stiffness $K_{d}$ values: 10, 15, and $20\;\frac{Nm}{deg}$ over the two compliant terrains of 63 and $25\;\frac{kN}{m}$. Initially, the performance of the admittance controller in tracking the desired ankle quasi-stiffness is analyzed over the bilaterally compliant terrain. Then, the performance of the two controllers in terms of stability is evaluated using phase portraits and the two proposed stability measures, estimating the probability of falling. All post-processing was implemented and executed in MATLAB\textsuperscript{TM} version 9.7 (R2019b) (The MathWorks, Natick, MA USA). Data were statistically tested to determine significance using the Wilcoxon rank-sum test (non-parametric counterpart to the t-test) with an $\alpha$ value of 0.01~\cite{gibbons2014nonparametric}.


\subsection{Ankle Quasi-stiffness Tracking of Admittance Controller}
\label{results_admittance}

\begin{figure}[!t]
    \centering
    \includegraphics[width=0.49\textwidth]{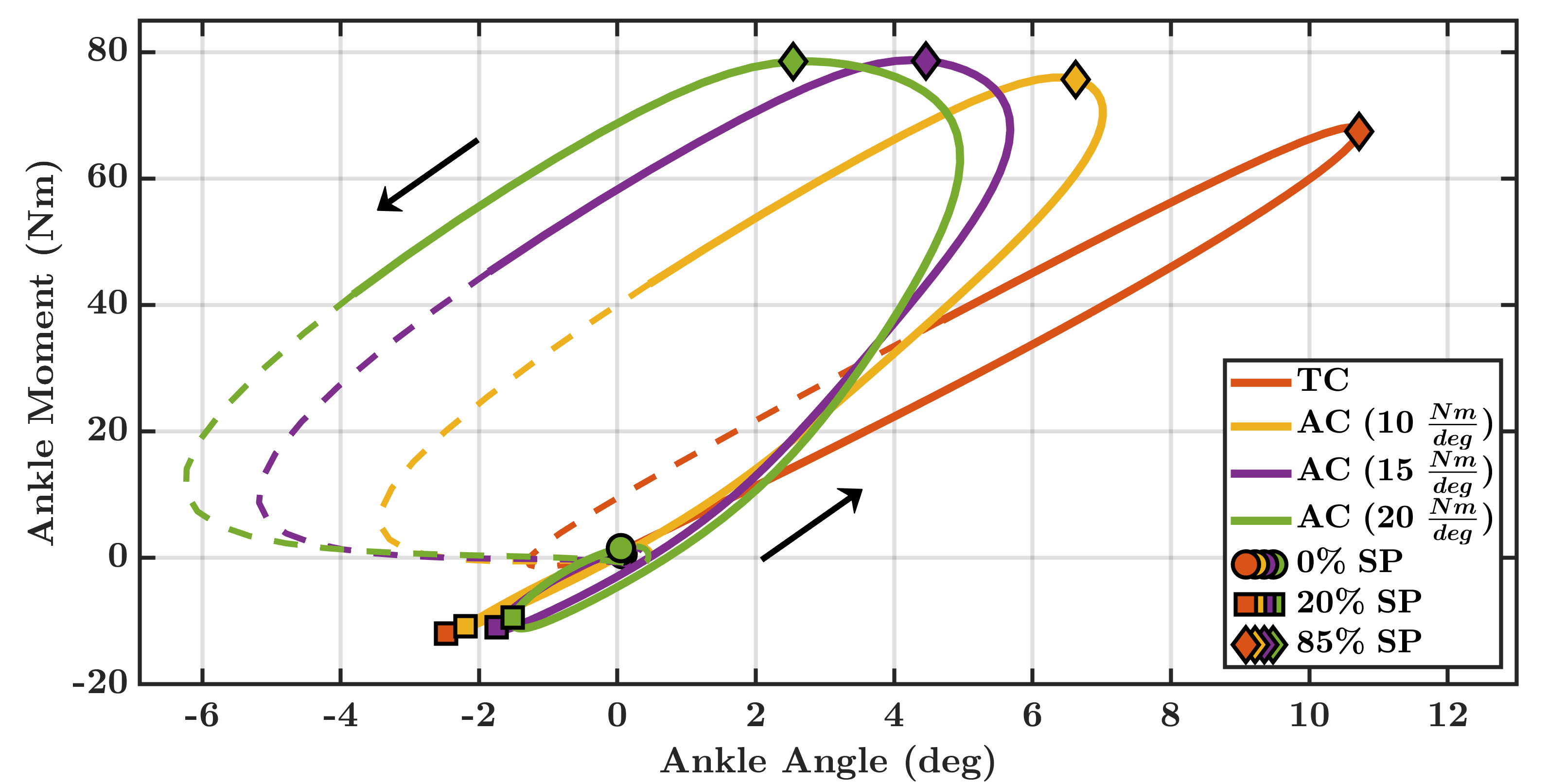}
    \vspace*{-1.5em}
    \caption{Moment-angle curves of the powered prosthesis ankle joint for the third subject walking over a bilaterally compliant terrain of $63\;\frac{kN}{m}$. The orange lines represent the \acf{TC}, while the yellow, purple, and green lines correspond to the \acf{AC} with desired stiffness values of $K_d=$ 10, 15, and $20\;\frac{Nm}{deg}$, respectively. Black arrows denote the increasing direction of the \acf{GC}. Solid and dashed lines signify the stance phase (0-60\% \ac{GC}) and the swing phase (60-100\% \ac{GC}), respectively. Circles ($\circ$), squares ($\square$), and diamonds ($\diamond$) mark the following time instances during \acf{SP}: foot-strike (0\% \ac{SP}), end of the loading response phase (20\% \ac{SP}), and the beginning of the pre-swing phase (85\% \ac{SP}). Dorsiflexion is indicated by a positive ankle angle.} 
    \label{fig:ankle_moment_curves}
\end{figure}

\begin{figure}[!t]
    \centering
    \includegraphics[width=0.49\textwidth]{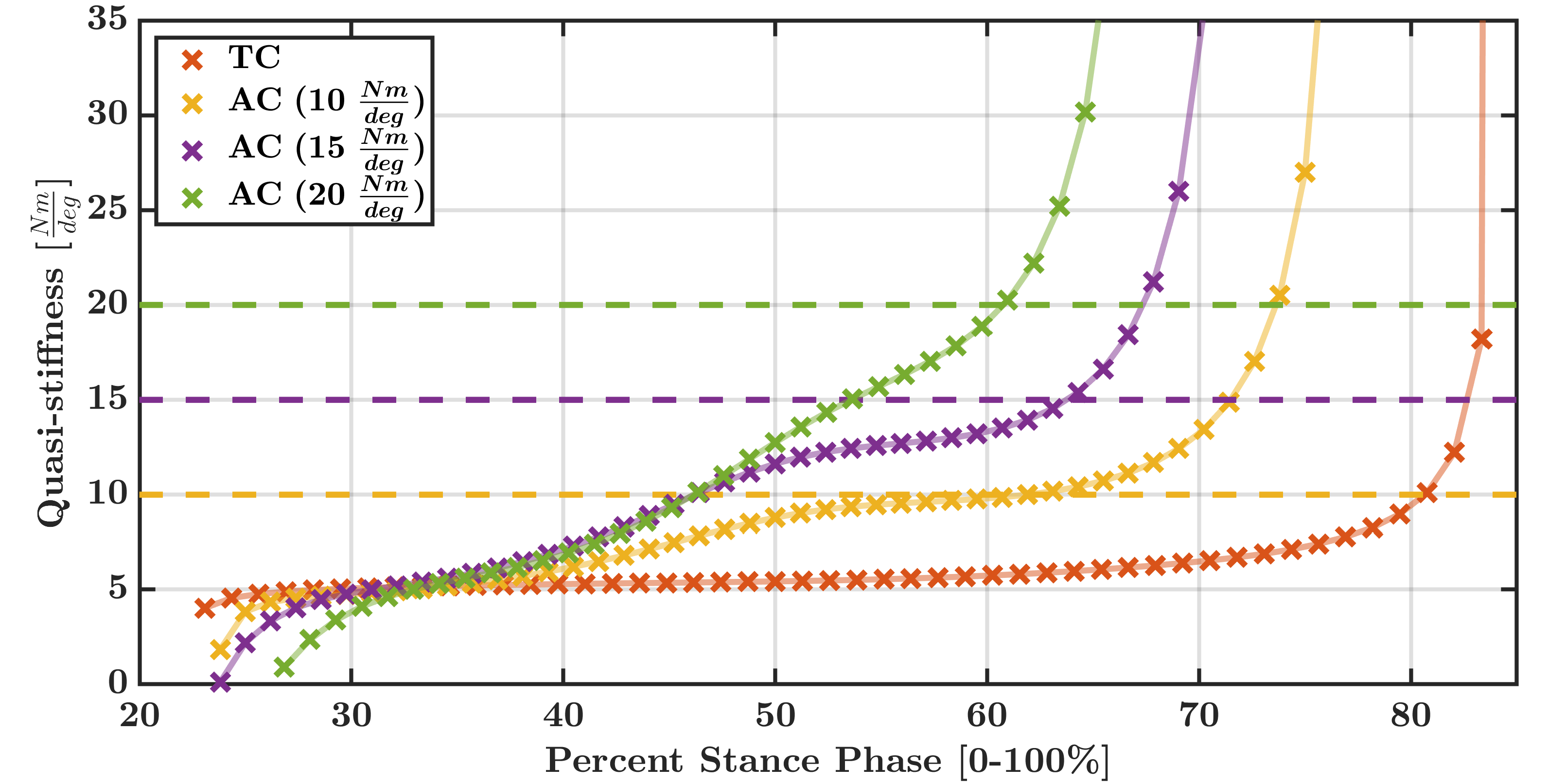}
    \vspace*{-1.5em}
    \caption{Ankle quasi-stiffness of the powered prosthesis ankle joint during stance phase for the third subject walking over a bilaterally compliant terrain of $63\;\frac{kN}{m}$. The orange line and crosses ($\times$) represent the \acf{TC}, while the yellow, purple, and green lines and crosses correspond to the \acf{AC} with desired stiffness values of $K_d=$ 10, 15, and $20\;\frac{Nm}{deg}$, respectively. Dashed lines indicate the desired stiffness values of $K_d=$ 10, 15, and $20\;\frac{Nm}{deg}$. No response is displayed during the loading response phase (0-20\% of the stance phase) and the pre-swing phase (85-100\% of the stance phase), due to high response variability arising from the reversal of the ankle moment-angle curve direction.} 
    \label{fig:quasi_stiffness}
\end{figure}

The performance of the admittance controller in tracking a desired quasi-stiffness was assessed while walking on compliant terrain for three desired stiffness $K_{d}$ levels: 10, 15 and $20\;\frac{Nm}{deg}$. Following the definition of~\cite{rouse2012difference}, the ankle quasi-stiffness was represented as the slope of the ankle joint moment-angle curve. Ankle quasi-stiffness was derived throughout the stance phase, defined as the first 60\% of the gait cycle starting at foot-strike~\cite{jacquelin2010gait}. The ankle angle and moment of the prosthesis, recorded at $100\;Hz$, were filtered using a $2^{nd}$ order, low-pass Butterworth filter (cut-off frequency of $5\;Hz$). As a representative example, the average moment-angle curves for the third subject during the trials over the first bilaterally compliant terrain of $63\;\frac{kN}{m}$ are depicted in Fig.~\ref{fig:ankle_moment_curves}. For each trial, the average moment-angle curve was derived by combining the average profiles of ankle moment and ankle angle across a total of 175 steps. For the sake of comparison, the corresponding average moment-angle curve for the \ac{TC} was similarly derived, where no control over the ankle quasi-stiffness was imposed. For the same trials, the response of the ankle quasi-stiffness during the stance phase is illustrated in Fig.~\ref{fig:quasi_stiffness}.

As depicted in Fig.~\ref{fig:ankle_moment_curves}, increasing the desired stiffness of the admittance controller results in a greater instantaneous slope of the moment-angle curve, and hence increased ankle quasi-stiffness. This is further supported by the observation that, when the admittance controller has a higher desired stiffness, there is a reduced angular deflection for the same applied load (moment). Based on these findings, it can be inferred that the ankle joint exhibits notably greater stiffness with all iterations of the admittance controller compared to the tibia controller.

The validity of the aforementioned observations is verified by the response of the ankle quasi-stiffness during the stance phase shown in Fig.~\ref{fig:quasi_stiffness}. In the case of the \ac{TC}, the prosthesis exhibits a quasi-stiffness that remains consistently near $5\;\frac{Nm}{deg}$ throughout the stance phase, even though no specific control was applied to regulate the quasi-stiffness. In comparison to the \ac{TC}, the \ac{AC} exhibits significantly higher quasi-stiffness values across the majority of the stance phase, while achieving the desired stiffness levels of $K_d=$ 10, 15, and $20\;\frac{Nm}{deg}$ around the onset of the terminal stance (60\% of the stance phase)~\cite{jacquelin2010gait}. Similarly to previous works, high variability of quasi-stiffness values was observed during the loading response (0-20\% of the stance phase) and the pre-swing (85-100\% of the stance phase) phases~\cite{rouse2014estimation}. This behavior is explained by the ankle angle reaching a plateau both before and after the reversal of the ankle moment-angle curve direction, as depicted in Fig.~\ref{fig:ankle_moment_curves}. Therefore, we decided to focus only on the response of the quasi-stiffness during the mid and terminal stance phases (20-85\% of the stance phase)~\cite{jacquelin2010gait}. In summary, the admittance controller succeeds in significantly increasing the prosthetic ankle's quasi-stiffness, reaching values nearly four times higher than those achieved by the tibia controller.

\subsection{Evaluation of the Tibia and Admittance Controllers}
\label{tibia_vs_admittance}

This section involves a comprehensive evaluation of both the admittance and tibia controllers over two bilaterally compliant terrains, in a total of eight walking trials for each of three subjects. The evaluation involves analyzing the phase portraits of the prosthetic leg, followed by assessing the maximum Lyapunov exponents and margins of stability, as detailed in Subsection~\ref{stability_measures}. As a reminder, left and right \ac{MOS} along the \ac{ML} and the \ac{AP} directions were evaluated for each gait cycle for a total of 175 steps across all trials. Then, average values and standard deviations were derived, and data were statistically tested to determine significance. In contrast, the short ($\lambda_{S}$) and long-term ($\lambda_{L}$) maximum Lyapunov exponents were evaluated for 25 overlapping windows of 150 strides for each trial. To facilitate the comparison of the proposed \ac{AC} relative to the standard \ac{TC}, the actual change:
\begin{equation}
    \Delta\lambda_{i} = \lambda_{i} - \lambda_{i}^{\text{TC}}, i \in \{S,L\},
\end{equation}
is visualized in subsequent figures for short-term $\Delta\lambda_{S}$ and long-term $\Delta\lambda_{L}$ maximum Lyapunov exponents across all controllers for each compliant terrain. As larger $\lambda_{S,L}$ values indicate an increased risk of falling, this suggests that $\Delta\lambda_{S,L}<0$, $\Delta\lambda_{S,L} \approx 0$, $\Delta\lambda_{S,L}>0$ indicate improved, similar, and deteriorated local dynamic stability, respectively, when compared to the \ac{TC} over each compliant terrain. Average values and standard deviations were derived across the 25 windows to show within-trial consistency. The results for both stability measures are also presented in tables as absolute numerical values, along with the average values across subjects.
\subsubsection{Phase Portraits}
\label{phase_portraits}
\begin{figure}[!t]
    \centering
    \includegraphics[width=0.49\textwidth]{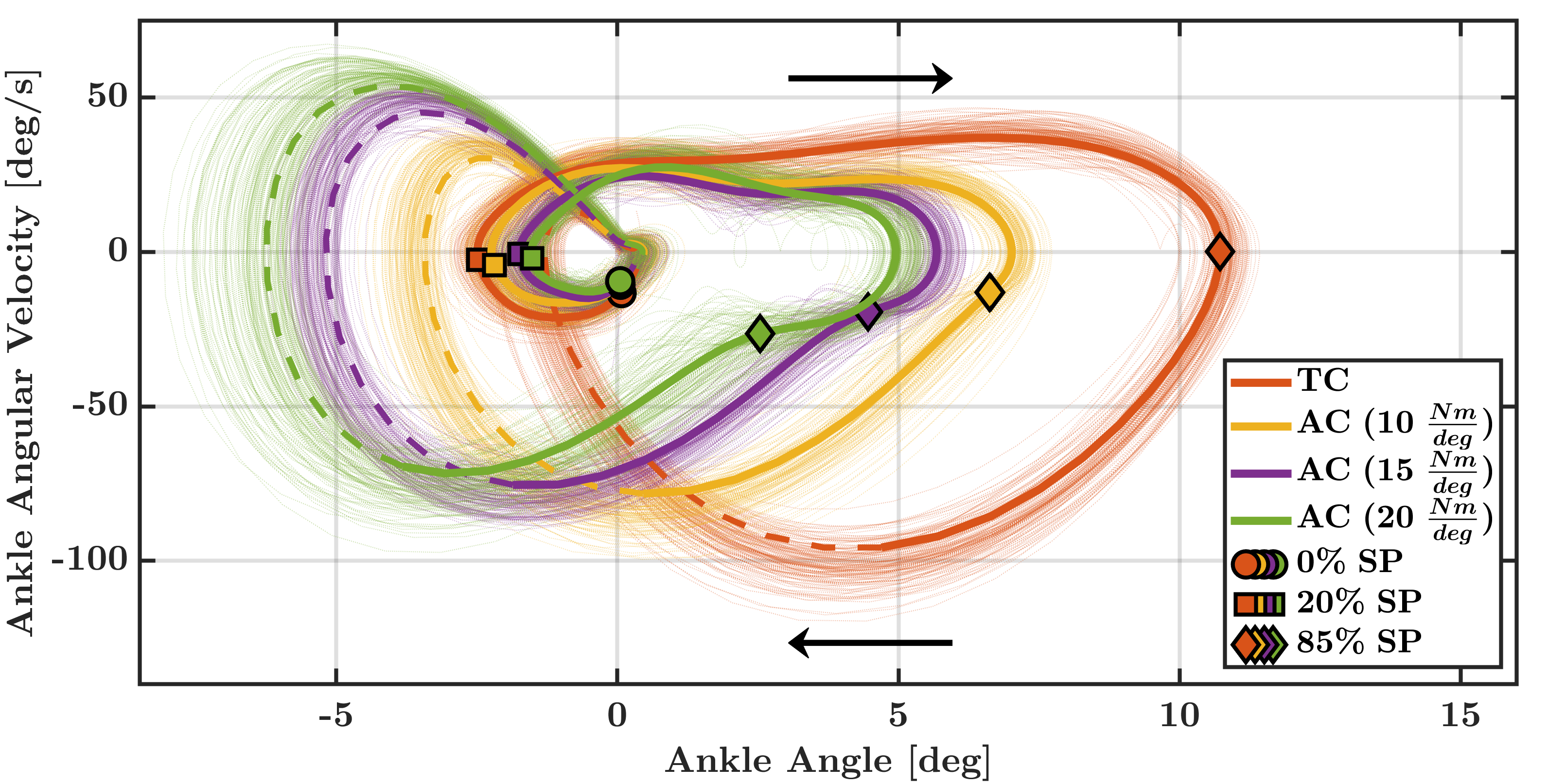}
    \vspace*{-1.5em}
    \caption{Phase portraits of the prosthetic leg (ankle angular velocity vs ankle angle) for the third subject walking over a bilaterally compliant terrain of $63\;\frac{kN}{m}$. The orange lines represent the \acf{TC}, while the yellow, purple, and green lines correspond to the \acf{AC} with desired stiffness values of $K_d=$ 10, 15, and $20\;\frac{Nm}{deg}$, respectively. Black arrows point towards the increasing direction of the \acf{GC}. Opaque lines denote the average phase portraits for each trial, with solid and dashed lines signifying the stance phase (0-60\% \ac{GC}) and the swing phase (60-100\% \ac{GC}), respectively. Semitransparent dotted lines illustrate the phase portraits of all gait cycles for each trial. Circles ($\circ$), squares ($\square$), and diamonds ($\diamond$) mark the time instances of the foot-strike (0\% of stance phase), the end of the loading response phase (20\% of stance phase), and the beginning of the pre-swing phase (85\% of stance phase). Dorsiflexion is indicated by a positive ankle angle.} 
    \label{fig:phase_plots}
\end{figure}

As a representative example, the phase portraits for the third subject during all four trials over the first bilaterally compliant terrain of $63\;\frac{kN}{m}$ are depicted in Fig.~\ref{fig:phase_plots}. It should be noted that similar responses were consistently observed across all three subjects for both bilaterally compliant terrains. For each trial, the phase portrait was derived by combining the profiles of ankle angular velocity and ankle angle across a total of 175 steps. Average phase portraits were also obtained for comparison purposes. The ankle angular velocities were first derived as the time derivatives of the ankle angles recorded at $100\;Hz$ on the prosthesis, and then they were filtered using a $2^{nd}$ order, low-pass Butterworth filter (cut-off frequency of $5\;Hz$). These phase portraits show that the prosthesis followed distinct stable periodic orbits for all controllers during the whole duration of the trials, enabling a stable steady-state gait pattern for the user.
\subsubsection{First Bilaterally Compliant Terrain of $63\;\frac{kN}{m}$}
\label{results_45k}
\begin{figure*}[t!] 
    \centering
  \subfloat[\scriptsize{Actual change $\Delta\lambda_{i} = \lambda_{i} - \lambda_{i}^{\text{TC}}, i \in \{S,L\}$ in short-term $\Delta\lambda_{S}$ (top) and long-term $\Delta\lambda_{L}$ (bottom) maximum Lyapunov exponents between all controllers and the \ac{TC} over compliant terrain. Positive and negative values indicate a less and more stable behavior than the \ac{TC} over compliant terrain, respectively.}]{%
    \label{fig:results_max_lyap_exp_45k}
       \includegraphics[width=0.48\linewidth]{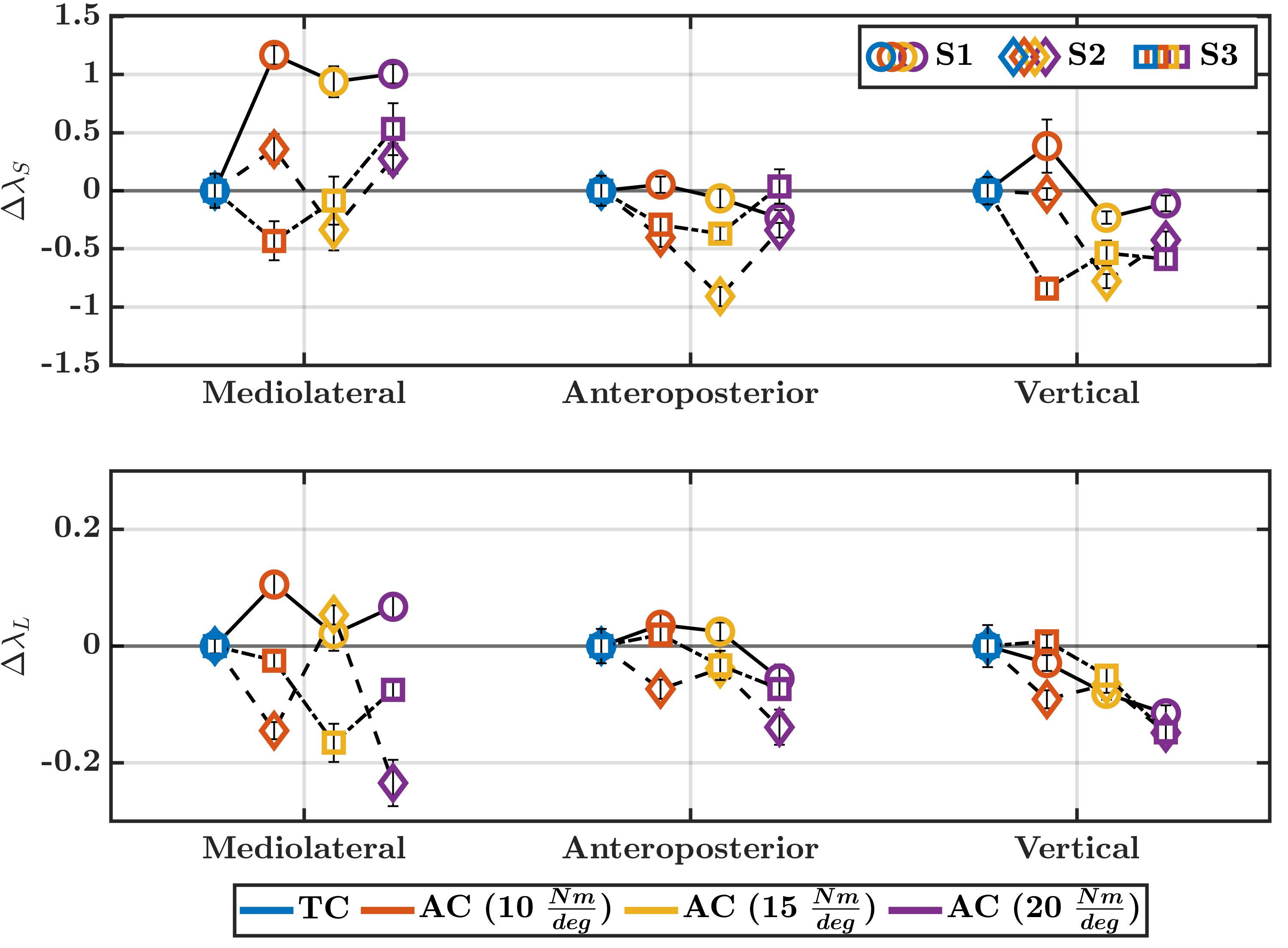}}
    \hfill
  \subfloat[\scriptsize{Mediolateral $\text{MOS}_{\text{ML}}$ (left) and anteroposterior $\text{MOS}_{\text{AP}}$ (right) margins of stability for the left-prosthetic and right-intact limb. Higher $\text{MOS}_{\text{ML}}$-$\text{MOS}_{\text{AP}}$ values indicate a more stable behavior.}]{%
        \includegraphics[width=0.48\linewidth]{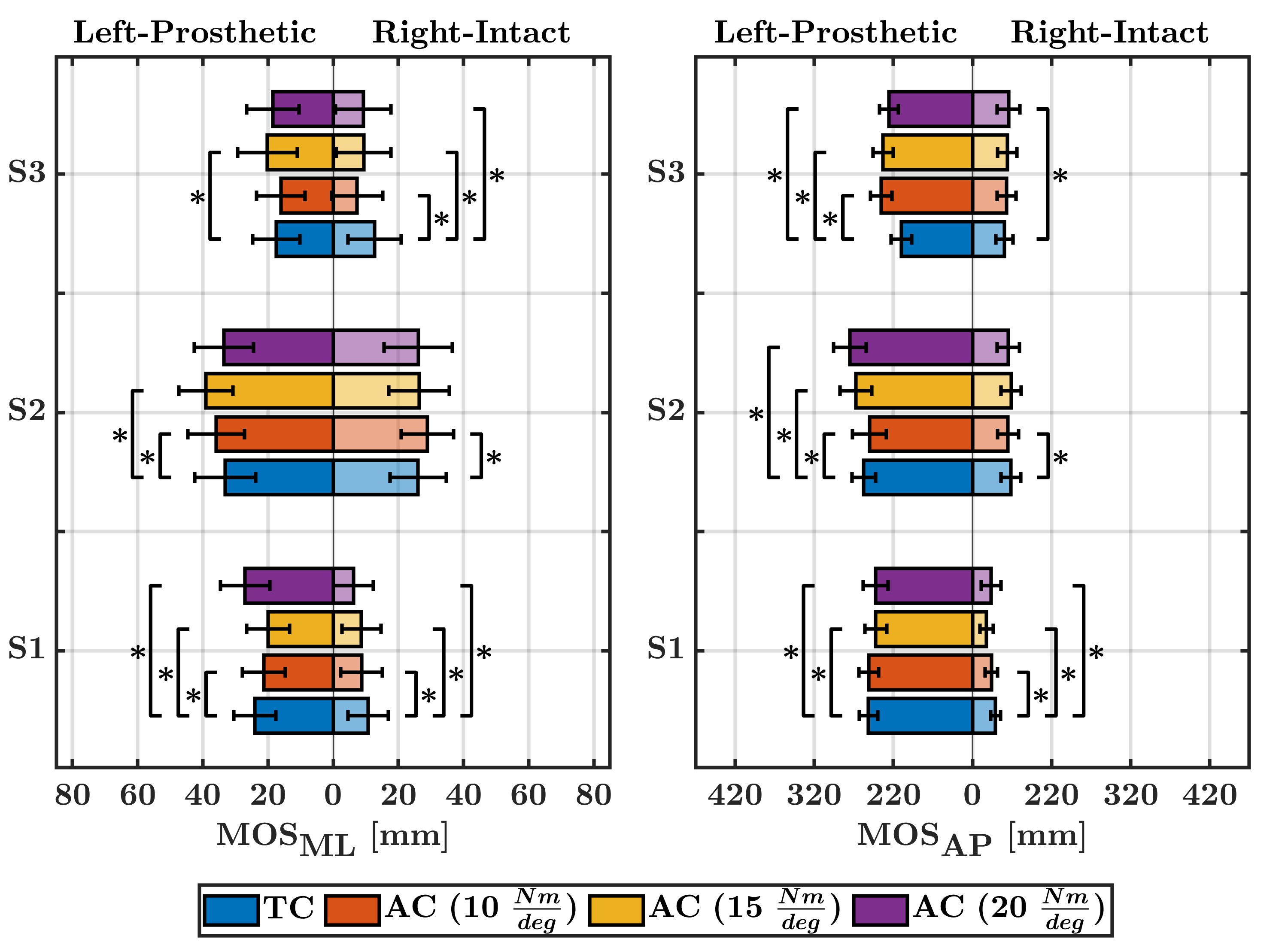}
        \label{fig:results_min_mos_45k}}
  \caption{Stability measures for walking trials over the first bilaterally compliant terrain of $63\;\frac{kN}{m}$ stiffness. Blue and orange markers and bars correspond to the \acf{TC} over rigid and compliant terrain, respectively, while yellow, purple, and green markers and bars denote the \acf{AC} over compliant terrain with desired stiffness values of 10, 15 and 20$\;\frac{Nm}{deg}$, respectively. In Fig.~\ref{fig:results_max_lyap_exp_45k}, circles ($\circ$), diamonds ($\diamond$), and squares ($\square$) represent the first, second, and third subjects, respectively. Similarly, solid, dashed, and dash-dotted black trendlines correspond to the first, second, and third subjects, respectively. Maximum Lyapunov exponents are expressed as ($\left<\ln\{d_{j}(i)\}\right>/\text{stride}$). Error bars represent within-trial standard deviations. In Fig.~\ref{fig:results_min_mos_45k}, asterisks (*) indicate significant differences between controllers ($p < 0.01$).} 
  \label{fig:res_stability_measures_45k} 
\end{figure*}

The stability evaluation results for the three subjects walking over both rigid and the first bilaterally compliant terrain with a ground stiffness of $63\;\frac{kN}{m}$ are depicted in Fig.~\ref{fig:res_stability_measures_45k} and Tables~\ref{table:short_long_term_values_63kNm}-\ref{table:left_right_mos_ml_ap_values_63kNm}. It should be noted that this ground stiffness resulted in a vertical deflection of both legs close to $10\;mm$. 
\begin{table*}[t!]
    \scriptsize
    \centering
    \textbf{Maximum Lyapunov Exponents for Walking Over 63 $\mathbf{\frac{kN}{m}}$ Compliant Terrain}\\[0.5em]  

    \resizebox{\textwidth}{!}{%
\begin{tabular}{|c||c|c|c|c||c|c|c|c||c|c|c|c|}
    \hline
    & \multicolumn{4}{c||}{\textbf{\acf{ML}} $\lambda_{S}$} & \multicolumn{4}{c||}{\textbf{\acf{AP}} $\lambda_{S}$} & \multicolumn{4}{c|}{\textbf{\acf{VT}} $\lambda_{S}$} \\
    \hline
    \textbf{Controller} & \textbf{Subject 1} & \textbf{Subject 2} & \textbf{Subject 3} & \textbf{Average} & \textbf{Subject 1} & \textbf{Subject 2} & \textbf{Subject 3} & \textbf{Average} & \textbf{Subject 1} & \textbf{Subject 2} & \textbf{Subject 3} & \textbf{Average} \\
    \hline
    TC                          & $7.13 \pm 0.09$   & $8.46 \pm 0.14$   & $7.93 \pm 0.15$   & $7.84 \pm 0.57$ & $7.27 \pm 0.07$   & $8.64 \pm 0.09$   & $7.07 \pm 0.13$   & $7.66 \pm 0.71$ & $5.65 \pm 0.05$  & $7.31 \pm 0.06$  & $5.66 \pm 0.12$ & $6.21 \pm 0.79$ \\ \hline
    AC ($10\;\frac{Nm}{deg}$)   & $8.30 \pm 0.09$   & $8.82 \pm 0.13$   & $\bm{7.49 \pm 0.17}$   & $8.20 \pm 0.57$ & $7.32 \pm 0.07$   & $\bm{8.24 \pm 0.09}$   & $\bm{6.77 \pm 0.10}$   & $\bm{7.44 \pm 0.61}$ & $6.04 \pm 0.23$  & $\bm{7.29 \pm 0.05}$  & $\bm{4.82 \pm 0.09}$ & $\bm{6.05 \pm 1.03}$ \\ \hline
    AC ($15\;\frac{Nm}{deg}$)   & $8.07 \pm 0.13$   & $\bm{8.13 \pm 0.18}$   & $\bm{7.84 \pm 0.21}$   & $8.01 \pm 0.21$ & $\bm{7.21 \pm 0.08}$   & $\bm{7.73 \pm 0.08}$   & $\bm{6.70 \pm 0.06}$   & $\bm{7.21 \pm 0.43}$ & $\bm{5.42 \pm 0.05}$  & $\bm{6.54 \pm 0.06}$   & $\bm{5.12 \pm 0.11}$ & $\bm{5.69 \pm 0.62}$ \\ \hline
    AC ($20\;\frac{Nm}{deg}$)   & $8.13 \pm 0.08$   & $8.74 \pm 0.13$   & $8.46 \pm 0.22$   & $8.44 \pm 0.29$ & $\bm{7.04 \pm 0.07}$   & $\bm{8.30 \pm 0.06}$   & $7.10 \pm 0.15$   & $\bm{7.48 \pm 0.59}$ & $\bm{5.54 \pm 0.07}$  & $\bm{6.89 \pm 0.08}$  & $\bm{5.07 \pm 0.07}$ & $\bm{5.83 \pm 0.78}$ \\ 
    \hline
    \end{tabular}
    }
\end{table*}
\begin{table*}[t!]
    \scriptsize
    \centering
    \resizebox{\textwidth}{!}{%
    \begin{tabular}{|c||c|c|c|c||c|c|c|c||c|c|c|c|}
        \hline
        & \multicolumn{4}{c||}{\textbf{\acf{ML}} $\lambda_{L}$} & \multicolumn{4}{c||}{\textbf{\acf{AP}} $\lambda_{L}$} & \multicolumn{4}{c|}{\textbf{\acf{VT}} $\lambda_{L}$} \\
        \hline
        \textbf{Controller} & \textbf{Subject 1} & \textbf{Subject 2} & \textbf{Subject 3} & \textbf{Average} & \textbf{Subject 1} & \textbf{Subject 2} & \textbf{Subject 3} & \textbf{Average} & \textbf{Subject 1} & \textbf{Subject 2} & \textbf{Subject 3} & \textbf{Average} \\
        \hline
        TC                          & $0.51 \pm 0.03$ & $0.44 \pm 0.02$ & $0.56 \pm 0.01$ & $0.50 \pm 0.05$ & $0.40 \pm 0.01$ & $0.29 \pm 0.02$ & $0.35 \pm 0.03$ & $0.34 \pm 0.05$ & $0.29 \pm 0.01$ & $0.21 \pm 0.01$ & $0.23 \pm 0.04$ & $0.25 \pm 0.05$ \\ \hline
        AC ($10\;\frac{Nm}{deg}$)   & $0.61 \pm 0.02$ & $\bm{0.29 \pm 0.02}$ & $\bm{0.53 \pm 0.02}$ & $\bm{0.48 \pm 0.14}$ & $0.43 \pm 0.02$ & $\bm{0.21 \pm 0.02}$ & $0.37 \pm 0.01$ & $0.34 \pm 0.10$ & $\bm{0.26 \pm 0.01}$ & $\bm{0.12 \pm 0.02}$ & $0.24 \pm 0.01$ & $\bm{0.21 \pm 0.06}$ \\ \hline
        AC ($15\;\frac{Nm}{deg}$)   & $0.53 \pm 0.03$ & $0.49 \pm 0.02$ & $\bm{0.39 \pm 0.03}$ & $\bm{0.47 \pm 0.06}$ & $0.42 \pm 0.02$ & $\bm{0.25 \pm 0.01}$ & $\bm{0.32 \pm 0.03}$ & $\bm{0.33 \pm 0.07}$ & $\bm{0.21 \pm 0.01}$ & $\bm{0.15 \pm 0.02}$ & $\bm{0.18 \pm 0.02}$ & $\bm{0.18 \pm 0.03}$ \\ \hline
        AC ($20\;\frac{Nm}{deg}$)   & $0.58 \pm 0.02$ & $\bm{0.20 \pm 0.04}$ & $\bm{0.48 \pm 0.01}$ & $\bm{0.42 \pm 0.16}$ & $\bm{0.34 \pm 0.02}$ & $\bm{0.15 \pm 0.03}$ & $\bm{0.28 \pm 0.02}$ & $\bm{0.26 \pm 0.08}$ & $\bm{0.18 \pm 0.01}$ & $\bm{0.07 \pm 0.02}$ & $\bm{0.09 \pm 0.02}$ & $\bm{0.11 \pm 0.05}$ \\ 
        \hline
    \end{tabular}
    }
\caption{Absolute short $\lambda_{S}$ (top) and long-term $\lambda_{L}$ (bottom) Lyapunov exponents for walking trials over the first bilaterally compliant terrain of $63\;\frac{kN}{m}$ stiffness across all subjects. Bold values for the admittance controllers indicate lower values compared to the tibia controller, and hence improved stability. Average columns report the average values across all subjects for each controller.}
    \label{table:short_long_term_values_63kNm}

\end{table*}

\begin{table*}[t!]
    \scriptsize
    \centering
    \textbf{ Margins of Stability in Mediolateral and Anteroposterior Directions Over 63 $\mathbf{\frac{kN}{m}}$ Compliant Terrain}\\[0.5em]  %
    \resizebox{\textwidth}{!}{%
    \begin{tabular}{|c||c|c|c|c||c|c|c|c||}
        \hline
        & \multicolumn{4}{c||}{\textbf{Left-Prosthetic $\text{MOS}_{\text{ML}}$}} & \multicolumn{4}{c||}{\textbf{Right-Intact $\text{MOS}_{\text{ML}}$}} \\
        \hline
        \textbf{Controller} & \textbf{Subject 1} & \textbf{Subject 2} & \textbf{Subject 3} & \textbf{Average} & \textbf{Subject 1} & \textbf{Subject 2} & \textbf{Subject 3} & \textbf{Average}\\
        \hline
        TC                          & $24.07 \pm 6.46$ & $33.14 \pm 9.31$ & $17.50 \pm 7.24$ & $24.90 \pm 10.06$ & $10.74 \pm 6.21$ & $26.07 \pm 8.64$ & $12.76 \pm 8.17$ & $16.52 \pm 10.30$\\ \hline
        AC ($10\;\frac{Nm}{deg}$)   & $21.30 \pm 6.62 (^{*})$ & $\bm{35.93 \pm 8.65 (^{*})}$ & $16.04 \pm 7.42$ & $24.43 \pm 11.34$ & $8.72 \pm 6.42 (^{*})$ & $\bm{28.96 \pm 8.01} (^{*})$ & $7.38 \pm 7.79 (^{*})$ & $15.02 \pm 12.36 (^{*})$ \\ \hline
        AC ($15\;\frac{Nm}{deg}$)   & $19.97 \pm 6.62 (^{*})$ & $\bm{39.05 \pm 8.32} (^{*})$ & $\bm{20.17 \pm 9.11} (^{*})$ & $\bm{26.40 \pm 12.05}$ & $8.64 \pm 6.00 (^{*})$ & $\bm{26.38 \pm 9.26}$ & $9.38 \pm 8.39 (^{*})$ & $14.80 \pm 11.45 (^{*})$ \\ \hline
        AC ($20\;\frac{Nm}{deg}$)   & $\bm{27.05 \pm 7.59} (^{*})$ & $\bm{33.53 \pm 9.15}$ & $\bm{18.51 \pm 8.06}$ & $\bm{26.36 \pm 10.31 (^{*})}$ & $6.20 \pm 6.09 (^{*})$ & $\bm{26.11 \pm 10.50}$ & $9.24 \pm 8.53 (^{*})$ & $13.85 \pm 12.24 (^{*})$ \\ 
        \hline
    \end{tabular}
    }
\end{table*}
\begin{table*}[t!]
    \scriptsize
    \centering
   
    \resizebox{\textwidth}{!}{%
    \begin{tabular}{|c||c|c|c|c||c|c|c|c||c|c|c|c||c|c|c|c||}
        \hline
        & \multicolumn{4}{c||}{\textbf{Left-Prosthetic $\text{MOS}_{\text{AP}}$}} & \multicolumn{4}{c||}{\textbf{Right-Intact $\text{MOS}_{\text{AP}}$}} \\
        \hline
        \textbf{Controller} & \textbf{Subject 1} & \textbf{Subject 2} & \textbf{Subject 3} & \textbf{Average} & \textbf{Subject 1} & \textbf{Subject 2} & \textbf{Subject 3} & \textbf{Average} \\
        \hline
        TC                          & $251.43 \pm 11.62$ & $257.33 \pm 15.01$ & $209.67 \pm 13.15$ & $239.47 \pm 25.06$ & $149.22 \pm 6.39$ & $168.51 \pm 12.30$ & $160.39 \pm 10.91$ & $159.37 \pm 12.88$ \\ \hline
        AC ($10\;\frac{Nm}{deg}$)   & $251.05 \pm 12.42$ & $250.16 \pm 21.26 (^{*})$ & $\bm{235.16 \pm 13.61} (^{*})$ & $\bm{245.46 \pm 17.76}$ & $144.10 \pm 7.86 (^{*})$ & $164.97 \pm 13.46 (^{*})$ & $\bm{162.98 \pm 11.87}$ & $157.35 \pm 14.70$ \\ \hline
        AC ($15\;\frac{Nm}{deg}$)   & $242.11 \pm 13.73 (^{*})$ & $\bm{267.21 \pm 20.01 (^{*})}$ & $\bm{233.02 \pm 12.72 (^{*})}$ & $\bm{247.45 \pm 21.41 (^{*})}$ & $137.87 \pm 8.21 (^{*})$ & $\bm{169.02 \pm 12.78}$ & $\bm{164.07 \pm 12.22}$ & $156.99 \pm 17.70$ \\ \hline
        AC ($20\;\frac{Nm}{deg}$)   & $242.19 \pm 15.79 (^{*})$ & $\bm{275.09 \pm 20.61} (^{*})$ & $\bm{225.59 \pm 12.13 (^{*})}$ & $\bm{247.62 \pm 26.39 (^{*})}$ & $143.63 \pm 12.37 (^{*})$ & $165.23 \pm 14.14$ & $\bm{165.75 \pm 14.40} (^{*})$ & $158.20 \pm 17.10$ \\ 
        \hline
    \end{tabular}
    }
    \caption{Mediolateral $\text{MOS}_{\text{ML}}$ (top) and anteroposterior $\text{MOS}_{\text{AP}}$ (bottom) margins of stability for the left-prosthetic and right-intact limb over the first bilaterally compliant terrain of $63\;\frac{kN}{m}$ stiffness across all subjects. Bold values for the admittance controllers indicate greater values compared to the tibia controller, and hence improved stability. Asterisks (*) indicate significant differences between each admittance controller and the tibia controller ($p < 0.01$). Average columns report the average values across all subjects for each controller.}
    \label{table:left_right_mos_ml_ap_values_63kNm}
\end{table*}

\paragraph{Short-term Maximum Lyapunov Exponents}
Compared to the \ac{TC}, the \ac{AC} overall led to lower ($\Delta\lambda_{S} < 0$) short-term maximum Lyapunov exponents ($\lambda_{S}$) for all subjects in the \ac{AP} and \ac{VT} directions over the $63\;\frac{kN}{m}$ compliant terrain (see Fig.~\ref{fig:results_max_lyap_exp_45k} and Table~\ref{table:short_long_term_values_63kNm}). The association of larger $\lambda_{S}$ values with increased probability of falling suggests improved walking stability with the proposed admittance controller in the \ac{AP} and \ac{VT} directions. For the \ac{ML} direction, although the \ac{AC} of $15\;\frac{Nm}{deg}$ led to lower $\lambda_{S}$ values for the second and third subjects ($\Delta\lambda_{S} < 0$), all \acp{AC} resulted in greater $\lambda_{S}$ values for the first subject ($\Delta\lambda_{S}>0$). Average responses across subjects support these findings, showing that the \ac{AC} improved walking stability in the \ac{AP} and \ac{VT} directions, while it deteriorated walking stability in the \ac{ML} direction (Table~\ref{table:short_long_term_values_63kNm}).

\paragraph{Long-term Maximum Lyapunov Exponents}
As depicted in the bottom sections of Fig.~\ref{fig:results_max_lyap_exp_45k} and Table~\ref{table:short_long_term_values_63kNm}, most versions of the \ac{AC} achieved lower ($\Delta\lambda_{L} < 0$) long-term maximum Lyapunov exponents ($\lambda_{L}$) than the \ac{TC} in all three velocity signals for the second and third subjects over the $63\;\frac{kN}{m}$ compliant terrain. As with $\lambda_{S}$, lower $\lambda_{L}$ values similarly suggest a reduced probability of falling and hence an improved walking stability. For the first subject, although lower $\lambda_{L}$ values are also observed for all three versions of the \ac{AC} in the \ac{VT} direction ($\Delta\lambda_{L} < 0$), greater values are seen for all three \acp{AC} in the \ac{ML} direction ($\Delta\lambda_{L} > 0$), while only the stiffest \ac{AC} ($20\;\frac{Nm}{deg}$) led to lower values in the \ac{AP} direction ($\Delta\lambda_{L} < 0$). Average responses across subjects support these findings, showing that the \ac{AC} improved walking stability in the \ac{AP} and \ac{VT} directions, while it deteriorated walking stability in the \ac{ML} direction (Table~\ref{table:short_long_term_values_63kNm}).

\paragraph{Mediolateral and Anteroposterior \ac{MOS}}
Figure~\ref{fig:results_min_mos_45k} and Table~\ref{table:left_right_mos_ml_ap_values_63kNm} show that all versions of the \ac{AC} achieved either significantly greater or no significantly different $\text{MOS}_{\text{ML}}$ and $\text{MOS}_{\text{AP}}$ compared to the \ac{TC} for the left-prosthetic side of the second and third subject over the $63\;\frac{kN}{m}$ compliant terrain. Given that greater $\text{MOS}$ values are associated with a lower probability of falling, this indicates that the \ac{AC} either improved or maintained the walking stability of the two subjects on the prosthesis side in both directions. For the right-intact leg of the second subject, all \acp{AC} led to either significantly greater or no significantly different $\text{MOS}_{\text{ML}}$ and $\text{MOS}_{\text{AP}}$ compared to the \ac{TC} in the majority of the cases. For the right side of the third subject, although all \acp{AC} led to either significantly greater or no significantly different $\text{MOS}_{\text{AP}}$ values compared to the \ac{TC}, significantly lower $\text{MOS}_{\text{ML}}$ were found with all versions of the \ac{AC}. For the first subject, all \acp{AC} led to significantly lower $\text{MOS}_{\text{ML}}$ and $\text{MOS}_{\text{AP}}$ values for both legs, except for the \acp{AC} of $20$ and $10\;\frac{Nm}{deg}$ that led to significantly increased $\text{MOS}_{\text{ML}}$ and non-significant different $\text{MOS}_{\text{AP}}$ values on the left side, respectively. Average responses across subjects support these findings, showing that the \ac{AC} improved walking stability in the \ac{ML} and \ac{AP} directions for the left-prosthetic side, while it deteriorated walking stability for the right-intact leg (Table~\ref{table:short_long_term_values_63kNm}). Finally, it should be noted that a consistent trend of asymmetry was evident across all subjects and controllers, with the margins of stability being greater for the prosthetic (left) leg in comparison to the intact (right) leg.

\subsubsection{Second Bilaterally Compliant Terrain of $25\;\frac{kN}{m}$}
\label{results_64k}

\begin{figure*}[t!] 
    \centering
  \subfloat[\scriptsize{Actual change $\Delta\lambda_{i} = \lambda_{i} - \lambda_{i}^{\text{TC}}, i \in \{S,L\}$ in short-term $\Delta\lambda_{S}$ (top) and long-term $\Delta\lambda_{L}$ (bottom) maximum Lyapunov exponents between all controllers and the \ac{TC} over compliant terrain. Positive and negative values indicate a less and more stable behavior than the \ac{TC} over compliant terrain, respectively.}]{%
        \label{fig:results_max_lyap_exp_64k}%
       \includegraphics[width=0.48\linewidth]{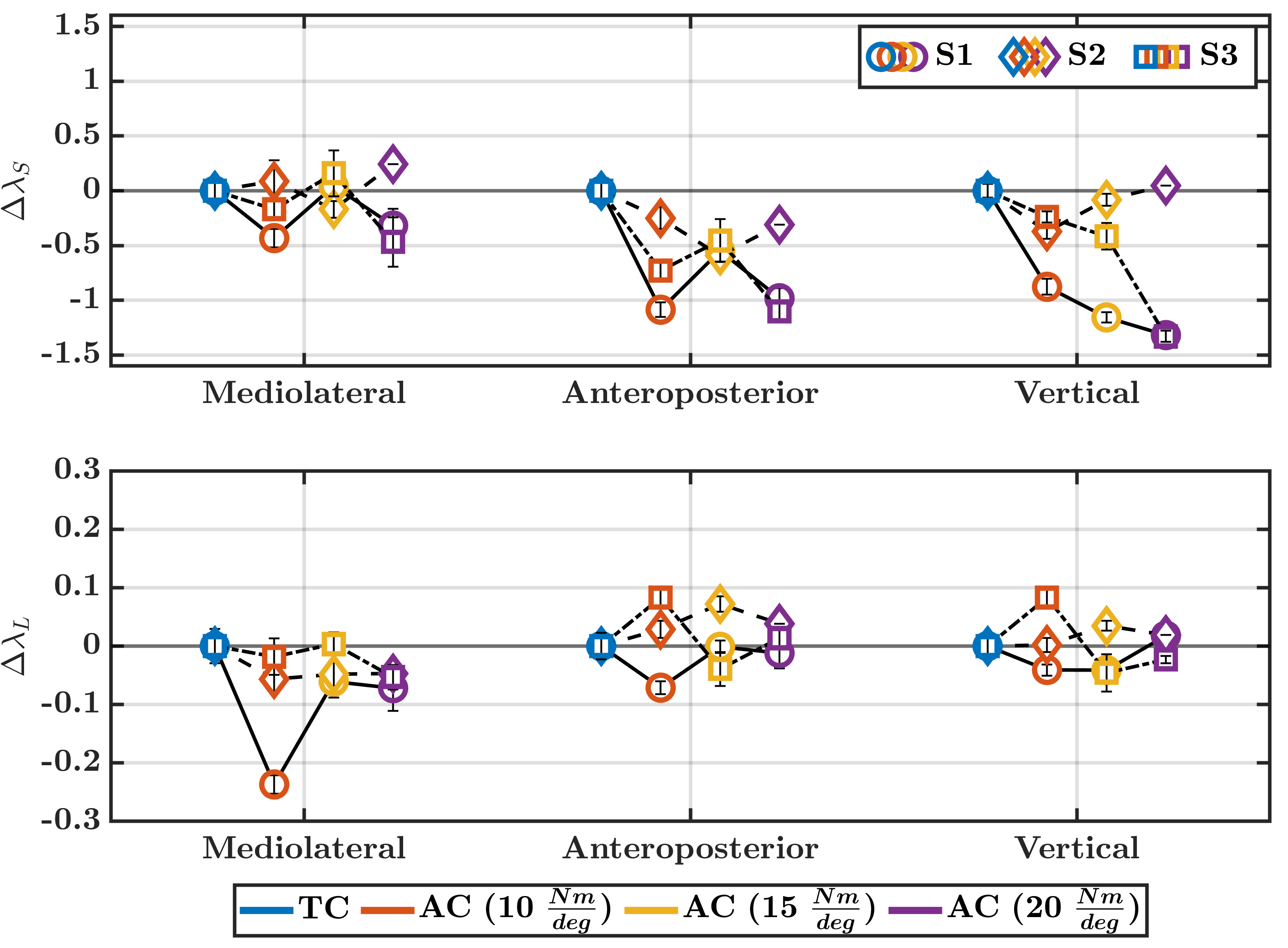}}
    \hfill
    \subfloat[\scriptsize{Mediolateral $\text{MOS}_{\text{ML}}$ (left) and anteroposterior $\text{MOS}_{\text{AP}}$ (right) margins of stability for the left-prosthetic and right-intact limb. Higher $\text{MOS}_{\text{ML}}$-$\text{MOS}_{\text{AP}}$ values indicate a more stable behavior.}]{%
  \label{fig:results_min_mos_64k}
        \includegraphics[width=0.48\linewidth]{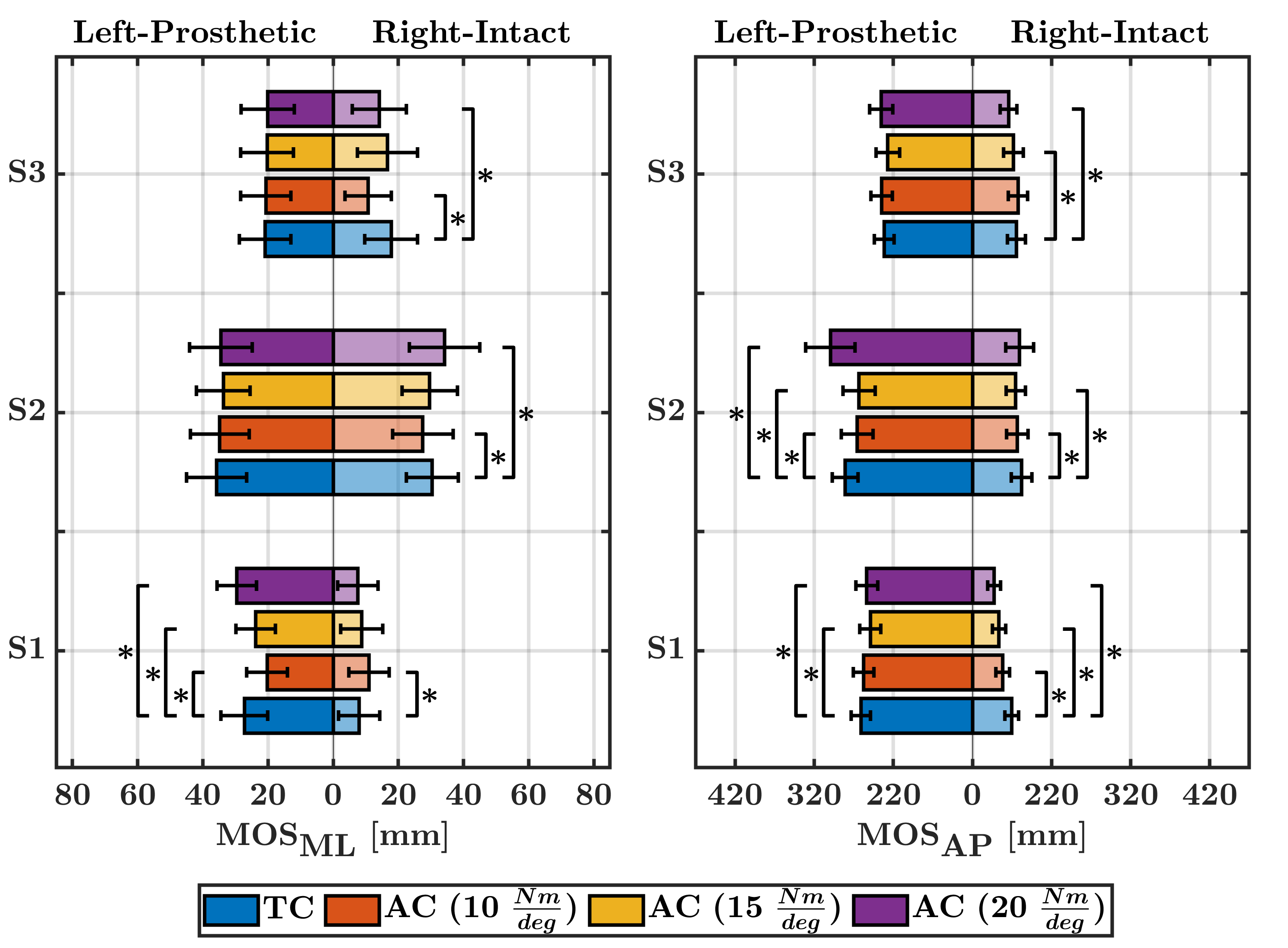}}
  \caption{Stability measures for walking trials over the second bilaterally compliant terrain of $25\;\frac{kN}{m}$ stiffness. Blue and orange markers and bars correspond to the \acf{TC} over rigid and compliant terrain, respectively, while yellow, purple, and green markers and bars denote the \acf{AC} over compliant terrain with desired stiffness values of 10, 15 and 20$\;\frac{Nm}{deg}$, respectively. In Fig.~\ref{fig:results_max_lyap_exp_64k}, circles ($\circ$), diamonds ($\diamond$), and squares ($\square$) represent the first, second, and third subjects, respectively. Similarly, solid, dashed, and dash-dotted black trendlines correspond to the first, second, and third subjects, respectively. Maximum Lyapunov exponents are expressed as ($\left<\ln\{d_{j}(i)\}\right>/\text{stride}$). Error bars represent within-trial standard deviations. In Fig.~\ref{fig:results_min_mos_64k}, asterisks (*) indicate significant differences between controllers ($p < 0.01$).}
  \label{fig:res_stability_measures_64k} 
\end{figure*}

The stability evaluation results for the three subjects walking over the second bilaterally compliant terrain with a ground stiffness of $25\;\frac{kN}{m}$ are depicted in Fig.~\ref{fig:res_stability_measures_64k} and Tables~\ref{table:short_long_term_stability_measures_25kN}-\ref{table:left_right_mos_ml_ap_values_25kNm}. Subjects experienced significant vertical deflections of both legs close to $20\;mm$ for this ground stiffness.

\begin{table*}[h!]
    \centering
        \textbf{Maximum Lyapunov Exponents for Walking Over 25 $\mathbf{\frac{kN}{m}}$ Compliant Terrain}\\[0.5em]  

    \resizebox{\textwidth}{!}{%
    \begin{tabular}{|c||c|c|c|c||c|c|c|c||c|c|c|c|}
        \hline
        & \multicolumn{4}{c||}{\textbf{\acf{ML}} $\lambda_{S}$} & \multicolumn{4}{c||}{\textbf{\acf{AP}} $\lambda_{S}$} & \multicolumn{4}{c|}{\textbf{\acf{VT}} $\lambda_{S}$} \\
        \hline
        \textbf{Controller} & \textbf{Subject 1} & \textbf{Subject 2} & \textbf{Subject 3} & \textbf{Average} & \textbf{Subject 1} & \textbf{Subject 2} & \textbf{Subject 3} & \textbf{Average} & \textbf{Subject 1} & \textbf{Subject 2} & \textbf{Subject 3} & \textbf{Average} \\
        \hline
        TC                          & $7.74 \pm 0.04$   & $8.74 \pm 0.09$   & $8.46 \pm 0.10$ & $8.11 \pm 0.38$ & $7.83 \pm 0.06$   & $8.85 \pm 0.11$   & $7.59 \pm 0.07$ & $7.74 \pm 0.22$ & $6.62 \pm 0.04$   & $7.72 \pm 0.08$   & $6.17 \pm 0.06$ & $6.42 \pm 0.30$ \\ \hline
        AC ($10\;\frac{Nm}{deg}$)   & $\bm{7.31 \pm 0.09}$   & $8.82 \pm 0.19$   & $\bm{8.29 \pm 0.10}$ & $\bm{7.81 \pm 0.51}$ & $\bm{6.75 \pm 0.07}$   & $\bm{8.60 \pm 0.10}$   & $\bm{6.86 \pm 0.07}$ & $\bm{6.84 \pm 0.29}$ & $\bm{5.74 \pm 0.07}$ & $\bm{7.35 \pm 0.07}$ & $\bm{5.94 \pm 0.05}$ & $\bm{5.87 \pm 0.26}$ \\ \hline
        AC ($15\;\frac{Nm}{deg}$)   & $7.78 \pm 0.14$   & $\bm{8.57 \pm 0.08}$   & $8.62 \pm 0.21$ & $8.21 \pm 0.46$ & $\bm{7.28 \pm 0.09}$   & $\bm{8.26 \pm 0.06}$   & $\bm{7.14 \pm 0.19}$ & $\bm{7.23 \pm 0.21}$ & $\bm{5.46 \pm 0.05}$ & $\bm{7.64 \pm 0.05}$ & $\bm{5.76 \pm 0.12}$ & $\bm{5.65 \pm 0.34}$ \\ \hline
        AC ($20\;\frac{Nm}{deg}$)   & $\bm{7.42 \pm 0.15}$   & $8.98 \pm 0$   & $\bm{7.99 \pm 0.23}$ & $\bm{7.73 \pm 0.39}$ & $\bm{6.85 \pm 0.09}$   & $\bm{8.54 \pm 0}$   & $\bm{6.49 \pm 0.08}$ & $\bm{6.71 \pm 0.33}$ & $\bm{5.30 \pm 0.07}$ & $7.77 \pm 0$ & $\bm{4.85 \pm 0.05}$ & $\bm{5.13 \pm 0.44}$ \\
        \hline
    \end{tabular}
    }
    
\end{table*}
\begin{table*}[h!]
    \centering
    \resizebox{\textwidth}{!}{%
    \begin{tabular}{|c||c|c|c|c||c|c|c|c||c|c|c|c|}
       \hline
        & \multicolumn{4}{c||}{\textbf{\acf{ML}} $\lambda_{L}$} & \multicolumn{4}{c||}{\textbf{\acf{AP}} $\lambda_{L}$} & \multicolumn{4}{c|}{\textbf{\acf{VT}} $\lambda_{L}$} \\
        \hline
        \textbf{Controller} & \textbf{Subject 1} & \textbf{Subject 2} & \textbf{Subject 3} & \textbf{Average} & \textbf{Subject 1} & \textbf{Subject 2} & \textbf{Subject 3} & \textbf{Average} & \textbf{Subject 1} & \textbf{Subject 2} & \textbf{Subject 3} & \textbf{Average} \\
        \hline
        TC                          & $0.60 \pm 0.01$ & $0.50 \pm 0.03$ & $0.56 \pm 0.02$ & $0.58 \pm 0.03$ & $0.40 \pm 0.02$ & $0.23 \pm 0.02$ & $0.31 \pm 0.02$ & $0.35 \pm 0.05$ & $0.22 \pm 0.01$ & $0.11 \pm 0.01$ & $0.14 \pm 0.01$ & $0.18 \pm 0.04$ \\ \hline
        AC ($10\;\frac{Nm}{deg}$)   & $\bm{0.36 \pm 0.02}$ & $\bm{0.45 \pm 0.02}$ & $\bm{0.54 \pm 0.03}$ & $\bm{0.45 \pm 0.09}$ & $\bm{0.33 \pm 0.01}$ & $0.26 \pm 0.02$ & $0.39 \pm 0.02$ & $0.36 \pm 0.04$ & $\bm{0.18 \pm 0.01}$ & $0.11 \pm 0.01$ & $0.23 \pm 0.01$ & $0.20 \pm 0.03$ \\ \hline
        AC ($15\;\frac{Nm}{deg}$)   & $\bm{0.54 \pm 0.03}$ & $\bm{0.46 \pm 0.02}$ & $0.56 \pm 0.02$ & $\bm{0.55 \pm 0.03}$ & $0.40 \pm 0.01$ & $0.30 \pm 0.01$ & $\bm{0.27 \pm 0.03}$ & $\bm{0.33 \pm 0.07}$ & $\bm{0.18 \pm 0.01}$ & $0.14 \pm 0.01$ & $\bm{0.10 \pm 0.03}$ & $\bm{0.14 \pm 0.05}$ \\ \hline
        AC ($20\;\frac{Nm}{deg}$)   & $\bm{0.53 \pm 0.04}$ & $\bm{0.46 \pm 0}$   & $\bm{0.51 \pm 0.02}$ & $\bm{0.52 \pm 0.03}$ & $\bm{0.39 \pm 0.03}$ & $0.27 \pm 0$ & $0.32 \pm 0.02$ & $0.35 \pm 0.04$ & $0.24 \pm 0.01$ & $0.13 \pm 0$ & $\bm{0.12 \pm 0.01}$ & $0.18 \pm 0.06$ \\ 
        \hline
    \end{tabular}
    }
    \caption{Absolute short $\lambda_{S}$ (top) and long-term $\lambda_{L}$ (bottom) Lyapunov exponents for walking trials over the second bilaterally compliant terrain of $25\;\frac{kN}{m}$ stiffness across all subjects. Bold values for the admittance controllers indicate lower values compared to the tibia controller, and hence improved stability. Average columns report the average values across all subjects for each controller.}
\label{table:short_long_term_stability_measures_25kN}
\end{table*}

\begin{table*}[h!]
    \scriptsize
    \centering
        \textbf{ Margins of Stability in Mediolateral and Anteroposterior Directions Over 25 $\mathbf{\frac{kN}{m}}$ Compliant Terrain}\\[0.5em]  %

    \resizebox{\textwidth}{!}{%
    \begin{tabular}{|c||c|c|c|c||c|c|c|c||c|c|c|c||c|c|c|c||}
        \hline
        & \multicolumn{4}{c||}{\textbf{Left-Prosthetic $\text{MOS}_{\text{ML}}$}} & \multicolumn{4}{c||}{\textbf{Right-Intact $\text{MOS}_{\text{ML}}$}} \\
        \hline
        \textbf{Controller} & \textbf{Subject 1} & \textbf{Subject 2} & \textbf{Subject 3} & \textbf{Average} & \textbf{Subject 1} & \textbf{Subject 2} & \textbf{Subject 3} & \textbf{Average} \\
        \hline
        TC                          & $27.25 \pm 7.15$ & $35.78 \pm 9.22$ & $20.90 \pm 7.94$ & $27.97 \pm 10.17$ & $7.93 \pm 6.32$ & $30.40 \pm 7.98$ & $17.82 \pm 8.09$ & $18.72 \pm 11.87$ \\ \hline
        AC ($10\;\frac{Nm}{deg}$)   & $20.27 \pm 6.21 (^{*})$ & $34.80 \pm 9.04$ & $20.68 \pm 7.75$ & $25.25 \pm 10.28 (^{*})$ & $\bm{11.04 \pm 6.20} (^{*})$ & $27.51 \pm 9.29 (^{*})$ & $10.79 \pm 7.12 (^{*})$ & $16.45 \pm 10.93 (^{*})$ \\ \hline
        AC ($15\;\frac{Nm}{deg}$)   & $23.77 \pm 6.07 (^{*})$ & $33.71 \pm 8.22$ & $20.26 \pm 8.12$ & $25.91 \pm 9.44 (^{*})$ & $\bm{8.79 \pm 6.44}$ & $29.61 \pm 8.52$ & $16.67 \pm 9.24$ & $18.35 \pm 11.83$ \\ \hline
        AC ($20\;\frac{Nm}{deg}$)   & $\bm{29.59 \pm 6.12} (^{*})$ & $34.47 \pm 9.67$ & $20.12 \pm 8.15$ & $27.76 \pm 9.98$ & $7.56 \pm 6.18$ & $\bm{34.18 \pm 10.79} (^{*})$ & $14.18 \pm 8.30 (^{*})$ & $17.92 \pm 13.95$ \\ 
        \hline
    \end{tabular}
    }
\end{table*}
\begin{table*}[h!]
    \scriptsize
    \centering
    \resizebox{\textwidth}{!}{%
    \begin{tabular}{|c||c|c|c|c||c|c|c|c||c|c|c|c||c|c|c|c||}
        \hline
        & \multicolumn{4}{c||}{\textbf{Left-Prosthetic $\text{MOS}_{\text{AP}}$}} & \multicolumn{4}{c||}{\textbf{Right-Intact $\text{MOS}_{\text{AP}}$}} \\
        \hline
        \textbf{Controller} & \textbf{Subject 1} & \textbf{Subject 2} & \textbf{Subject 3} & \textbf{Average} & \textbf{Subject 1} & \textbf{Subject 2} & \textbf{Subject 3} & \textbf{Average} \\
        \hline
        TC                          & $261.00 \pm 12.07$ & $280.98 \pm 16.36$ & $231.54 \pm 12.27$ & $257.84 \pm 24.50$ & $169.77 \pm 8.56$ & $182.09 \pm 12.90$ & $175.77 \pm 11.49$ & $175.87 \pm 12.19$ \\ \hline
        AC ($10\;\frac{Nm}{deg}$)   & $257.69 \pm 13.10$ & $265.91 \pm 20.05 (^{*})$ & $\bm{234.89 \pm 13.62}$ & $252.83 \pm 20.60 (^{*})$ & $158.10 \pm 8.74 (^{*})$ & $176.63 \pm 13.65 (^{*})$ & $\bm{177.53 \pm 12.19}$ & $170.75 \pm 14.73 (^{*})$ \\ \hline
        AC ($15\;\frac{Nm}{deg}$)   & $249.13 \pm 13.40 (^{*})$ & $263.35 \pm 20.37 (^{*})$ & $226.94 \pm 14.99$ & $246.47 \pm 22.29 (^{*})$ & $153.56 \pm 8.33 (^{*})$ & $174.64 \pm 12.29 (^{*})$ & $171.79 \pm 12.56 (^{*})$ & $166.67 \pm 14.59 (^{*})$ \\ \hline
        AC ($20\;\frac{Nm}{deg}$)   & $253.61 \pm 13.70 (^{*})$ & $\bm{299.58 \pm 31.13} (^{*})$ & $\bm{235.08 \pm 14.66}$ & $\bm{261.05 \pm 33.71}$ & $147.42 \pm 8.28 (^{*})$ & $179.57 \pm 17.57$ & $165.59 \pm 10.65 (^{*})$ & $163.48 \pm 18.06 (^{*})$ \\ 
        \hline
    \end{tabular}
    }
    \caption{Mediolateral $\text{MOS}_{\text{ML}}$ (top) and anteroposterior $\text{MOS}_{\text{AP}}$ (bottom) margins of stability for the left-prosthetic and right-intact limb over the second bilaterally compliant terrain of $25\;\frac{kN}{m}$ stiffness across all subjects. Bold values for the admittance controllers indicate greater values compared to the tibia controller, and hence improved stability. Asterisks (*) indicate significant differences between each admittance controller and the tibia controller ($p < 0.01$). Average columns report the average values across all subjects for each controller.}
        \label{table:left_right_mos_ml_ap_values_25kNm}

\end{table*}

\paragraph{Short-term Maximum Lyapunov Exponents}
The top sections of Fig.~\ref{fig:results_max_lyap_exp_64k} and Table~\ref{table:short_long_term_stability_measures_25kN} present that almost all versions of the \acf{AC} exhibited lower ($\Delta\lambda_{S}<0$) short-term maximum Lyapunov exponents than the \acf{TC} over the $25\;\frac{kN}{m}$ compliant terrain for all three subjects in all three \ac{CoM} velocity signals. The correlation between increased probability of falling and greater $\lambda_{S}$ values indicates that the walking stability improved with the proposed admittance controller. The only exceptions to the aforementioned response were found in the \ac{ML} direction, where greater values ($\Delta\lambda_{S}>0$) than the \ac{TC} over compliant terrain were identified for the second subject with an AC of $20\;\frac{Nm}{deg}$ and the third subject with an AC of $15\;\frac{Nm}{deg}$. Average responses across subjects support these findings, showing that the \ac{AC} improved walking stability in all three directions (see Table~\ref{table:short_long_term_stability_measures_25kN}).

\paragraph{Long-term Maximum Lyapunov Exponents}
The bottom sections of Fig.~\ref{fig:results_max_lyap_exp_64k} and Table~\ref{table:short_long_term_stability_measures_25kN} demonstrate that all versions of the \ac{AC} achieved similar or significantly lower ($\Delta\lambda_{L} < 0$) long-term maximum Lyapunov exponent values than the \ac{TC} over the $25\;\frac{kN}{m}$ compliant terrain in the \ac{ML} direction for all subjects. Similarly to the relationship seen with $\lambda_{L}$, lower $\lambda_{L}$ values imply a reduced probability of falling and hence an improved walking stability with the proposed admittance controller along the \ac{ML} direction. In the \ac{AP} and \ac{VT} directions, a varying behavior was observed across subjects, with the \ac{AC} leading to lower or similar $\lambda_{L}$ for the first subject ($\Delta\lambda_{L} < 0$), and greater or similar values for the second and third subjects ($\Delta\lambda_{L} > 0$). Nevertheless, it should be noted that as the desired stiffness of the \ac{AC} increased, the \ac{AC} exhibited similar $\lambda_{L}$ values to the \ac{TC} for all subjects in both directions ($\Delta\lambda_{L}\approx 0$). Average responses across subjects support these findings, showing that the \ac{AC} improved walking stability in the \ac{ML} direction and deteriorated it in the \ac{AP} and \ac{VT} directions (Table~\ref{table:short_long_term_stability_measures_25kN}).

\paragraph{Mediolateral and Anteroposterior \ac{MOS}}
Figure~\ref{fig:results_min_mos_64k} and Table~\ref{table:left_right_mos_ml_ap_values_25kNm} show no significant difference between the \ac{TC} and most versions of the \ac{AC} over the $25\;\frac{kN}{m}$ compliant terrain in the \ac{ML} and \ac{AP} directions for the left-prosthetic side of the second and third subject. Given that larger $\text{MOS}$ values are indicative of a lower probability of falling, this suggests that over this compliant terrain, the \ac{AC} largely maintained similar walking stability to the \ac{TC} in both directions for the prosthesis side of the two subjects. The observed pattern was largely consistent, with exceptions noted only for the second subject in the \ac{AP} direction. On the right-intact side of the second and third subjects, generally all \acp{AC} led to either significantly lower or non-significantly different $\text{MOS}_{\text{ML}}$ and $\text{MOS}_{\text{AP}}$ compared to the \ac{TC}, except for the significantly greater $\text{MOS}_{\text{ML}}$ found for the second subject with the \ac{AC} of $20\;\frac{Nm}{deg}$. 
For the first subject, significantly lower or non-significantly different margins of stability were identified for most versions of the \ac{AC} in both directions for both legs. Average responses across subjects indicate that overall the \ac{AC} deteriorated walking stability in both directions for both sides (Table~\ref{table:left_right_mos_ml_ap_values_25kNm}). Similarly to the first compliant terrain of $63\;\frac{kN}{m}$, the same left-right asymmetry was observed in this compliant terrain as well, with the prosthetic-left leg exhibiting greater margins of stability compared to the right-intact leg. 

\section{Discussion}
\label{discussion}
This work presented a novel admittance controller for improving the walking stability of ankle-foot prostheses across various compliant terrains. The proposed controller was evaluated with three healthy non-disabled subjects walking over two bilaterally compliant surfaces with ground stiffness values of $63$ and $25\;\frac{kN}{m}$. The results of this evaluation were presented in the previous section (\ref{results}), where the proposed controller was evaluated in terms of tracking a desired quasi-stiffness, as well as in terms of stability and compared to a standard (\ac{TC}) controller. This section provides a comprehensive summary of the key findings and the significant contributions of this work to the field, while also addressing its limitations and suggesting promising avenues for future investigation.

\subsection{Summary of Main Results}


First, the proposed \ac{AC} was capable of substantially increasing the ankle quasi-stiffness of the prosthesis, achieving approximately four times the stiffness exhibited by the \ac{TC} (see Section \ref{results_admittance}). Specifically, for each desired stiffness value $K_d=$ 10, 15, and $20\;\frac{Nm}{deg}$, the \ac{AC} exhibited distinct and continuously increasing quasi-stiffness profiles during the stance phase, while reaching the desired stiffness levels around the onset of the terminal stance (60\% of the stance phase)~\cite{jacquelin2010gait}. 

The performance of the \ac{AC} was compared to that of the \ac{TC} in terms of stability, using phase portraits, maximum Lyapunov exponents, and margins of stability. The phase portraits indicated that the prosthesis was able to track distinct stable periodic orbits for all controllers during the whole duration of the eight trials, enabling a stable steady-state gait pattern for all three users over rigid and compliant terrains (see Section \ref{phase_portraits}).

For the first bilaterally compliant terrain ($63\;\frac{kN}{m}$), the \ac{AC} improved walking stability in all directions for the second and third subjects, based on both Lyapunov exponents and margins of stability. In contrast, stability for the first subject improved only in the \ac{VT} direction and declined in the others. On average, the \ac{AC} enhanced sagittal-plane stability but reduced mediolateral stability, likely due to its control focus on prosthetic ankle plantar/dorsiflexion.

For the second bilaterally compliant terrain ($25\;\frac{kN}{m}$), the \ac{AC} generally improved or maintained local dynamic stability across all subjects, as indicated by the maximum Lyapunov exponents. Margins of stability showed similar or reduced stability compared to the \ac{TC}, particularly on the right-intact leg, with the first subject experiencing the most consistent reductions. Despite narrower margins, local stability was not negatively impacted, suggesting the margins remained sufficient. On average, the \ac{AC} either improved or maintained local dynamic stability, at the expense of narrower margins of stability on both sides. Overall, the \ac{AC} effectively maintained or improved local dynamic stability for all subjects, even under lower-stiffness conditions.

\subsection{Effect of Ankle Quasi-Stiffness on Gait Stability Across Compliant Terrains}

The improved local dynamic stability with the \ac{AC} compared to the \ac{TC} highlights the importance of adjusting ankle quasi-stiffness based on ground compliance. On the $63;\frac{kN}{m}$ terrain, stiffer \acp{AC} consistently improved stability for the second and third subjects, with Lyapunov exponents showing a trend of increased stability with higher ankle stiffness. A similar pattern was observed on the $25\;\frac{kN}{m}$ terrain, primarily in short-term stability, though overall improvements were less pronounced. These findings align with prior work showing that increased ankle or leg stiffness enhances stability in both human and bipedal robot models~\cite{karakasis2023adjusting,karakasis2022robust,karakasis2024energy}, and are consistent with biomechanics literature indicating that humans increase leg stiffness in response to softer surfaces~\cite{farley1998mechanism,xie2021compliance}.

\subsection{User Preferences and Perceived Comfort Across Terrain and Controller Conditions}

User preference for ankle quasi-stiffness appeared to influence perceived walking stability~\cite{clites2021understanding}. After each trial, participants were asked to rate comfort and compare the controllers. During initial training on rigid terrain, all subjects preferred the \ac{TC}, finding the \acp{AC} too spring-like and stiff. However, during longer trials on compliant terrain, subjects 1 and 3 favored the lower-stiffness \acp{AC} over the \ac{TC}, while subject 2 preferred the \ac{TC} on the stiffer terrain ($63\;\frac{kN}{m}$) and the $10\;\frac{Nm}{deg}$ \ac{AC} on the softer terrain ($25\;\frac{kN}{m}$). They noted that increased quasi-stiffness, though excessive on firmer ground, enhanced push-off and balance on softer surfaces.

Optimal ankle quasi-stiffness should account for ground stiffness, walking stability, and user comfort. A recent study identified user-preferred stiffness levels for variable-stiffness prostheses and linked these preferences to biomechanical and behavioral factors~\cite{clites2021understanding}. Building on this, the same group developed an offline machine learning model that accurately predicts a user's preferred ankle stiffness~\cite{shetty2022data}. In parallel, a pattern-recognition algorithm has been proposed to detect upcoming surface stiffness changes in real time using EMG and kinematic data~\cite{angelidou2023intuitive,angelidou2023predicting}. Together with the findings of this study, these advances support the development of adaptive control frameworks that adjust ankle stiffness in real time based on user input, stability needs, and terrain properties.


\subsection{Limitations of the Study and Future Work}

The main limitation of this study is that the prosthesis was tested on healthy, non-disabled individuals rather than participants with lower-limb amputation. Specifically, three subjects used a carbon fiber ankle bypass adapter mounted around their left shank to simulate prosthesis use. While this approach does not fully replicate the biomechanics of transtibial amputation, it is a widely accepted and commonly used method in prosthetics research~\cite{clark2022learning,karakasis2023adjusting,posh2023finite,gehlhar2023emulating}. The bypass adapter effectively isolates the biological ankle joint while preserving natural knee movement, making it a practical and well-established approximation. That said, it introduces some known limitations; most notably a mediolateral offset due to the prosthesis being mounted beside the intact limb (Fig.~\ref{fig:subject_prosthesis_vst}). This offset may influence step width and mediolateral stability, contributing to the left–right asymmetries observed in the margins of stability (Figs.~\ref{fig:res_stability_measures_45k}–\ref{fig:res_stability_measures_64k}). Similar asymmetries, however, have also been reported in both healthy individuals and those with transfemoral amputation~\cite{rosenblatt2010measures,hof2007control,young2012dynamic}. \P{Disentangling the effects of the adapter offset from natural gait variability is challenging and beyond the scope of this study. Nonetheless, by analyzing stability metrics separately for each side, we are able to draw meaningful conclusions about side-specific stability improvements across controllers and conditions.}

Additionally, limited exposure and training with the prosthesis may have influenced stability outcomes. While all subjects practiced with the device and controllers until comfortable, individuals with lower-limb amputation are likely to be more experienced and better adapted to prosthetic use. For this reason, future work will include testing with individuals with transtibial amputation to further validate the controller’s effectiveness over bilaterally compliant terrains.

While increasing ankle quasi-stiffness improved stability on the first compliant terrain, it did not significantly enhance gait stability on the softer $25\;\frac{kN}{m}$ surface, which is the lowest stiffness the \ac{VST 2} can achieve~\cite{chambers2025variable}. This may indicate that the terrain was too compliant for stiffness changes to have a meaningful effect. Moreover, the tested stiffness range ($10$–$20;\frac{Nm}{deg}$) aligns with values reported in both human and prosthetic studies~\cite{rouse2014estimation,farley1998mechanism,xie2021compliance,clites2021understanding}, making it unlikely that higher values would offer further benefit. However, other admittance controller parameters -- such as damping or equilibrium angle -- could also influence stability and warrant further exploration. These could be implemented as constants or time-varying functions, as seen in modern variable impedance controllers~\cite{mohammadi2019variable,best2023data,kumar2020impedance,lee2024towards,cortino2023data,reznick2024clinical}. While this study focused on isolating the stiffness effect, future work should investigate how tuning these additional parameters affects walking stability over compliant terrain.


While this study introduces a significant advancement in evaluating prosthesis control over compliant terrains, future research can build on this foundation by incorporating additional stability measures. The metrics used here were carefully selected for their relevance and reliability, but other measures, such as temporal gait symmetry, trunk pitch and roll angles, and angular velocity, have also been shown to effectively capture gait stability and postural control across various perturbation scenarios~\cite{honeycutt2016characteristics,owings2001mechanisms,marigold2005adapting,bruijn2013assessing}. Including such metrics in future studies could offer deeper insights into locomotion over compliant surfaces and further inform the development and evaluation of next-generation lower-limb prosthesis controllers.
\section{Conclusion}
\label{conclusion}


This work introduces a novel admittance controller specifically designed to enhance walking stability in ankle-foot prostheses across a range of compliant terrains. Its effectiveness was demonstrated through treadmill experiments with three healthy participants walking over two bilaterally compliant surfaces with stiffness values of $63$ and $25\;\frac{kN}{m}$. Compared to a standard phase-variable controller optimized for rigid ground, the proposed controller consistently improved local dynamic stability across all subjects and conditions. By addressing the critical challenge of maintaining stability on soft, real-world surfaces, this study marks a significant step toward more robust, adaptable prosthetic control. The findings lay the groundwork for advancing powered prostheses that can better support individuals with lower-limb amputation in navigating everyday environments with greater confidence and safety.

\balance

\bibliographystyle{IEEEtran}
\bibliography{references}


 



%

\end{document}